\documentclass[lettersize, journal]{IEEEtran}
\IEEEpubidadjcol

\IEEEoverridecommandlockouts                              %
\usepackage{soul}
\usepackage[normalem]{ulem}
\usepackage[hidelinks]{hyperref}
\usepackage{bookmark}
\usepackage[english]{babel}
\usepackage{mathtools}

\usepackage[dvipsnames]{xcolor}
\usepackage{amsfonts,amssymb,amsmath,color}
\usepackage{graphicx}
\usepackage{cite}
\usepackage{placeins}
\usepackage{graphicx}
\usepackage{algorithm}
\usepackage[noend]{algpseudocode}
\usepackage{todonotes}
\usepackage{balance}
\usepackage{tikz}
\usetikzlibrary{shapes,arrows,calc}
\usepackage{xpatch}
\usepackage{enumerate}

\allowdisplaybreaks
\onecolumn

\newcommand{\R}{\mathbb{R}}

\newcommand{\N}{\mathbb{N}}

\newcommand{\T}{^\top}

\newcommand{\until}[1]{\{1,\ldots,#1\}}

\newcommand{\subj}{\textnormal{subj.~to}}

\DeclareMathOperator{\diag}{diag}
\DeclareMathOperator{\col}{col}

\newtheorem{theorem}{Theorem}[section]

\newtheorem{assumption}[theorem]{Assumption}

\usepackage[dvipsnames]{xcolor}
\usepackage{tikz,pgfplots}
\pgfplotsset{compat=newest}

\newcommand\oprocendsymbol{\hbox{$\blacksquare$}}
\newcommand\oprocend{\relax\ifmmode\else\unskip\hfill\fi\oprocendsymbol}

\newcommand{\cA}{\mathcal A}

\newcommand{\cE}{\mathcal E}

\newcommand{\cG}{\mathcal G}

\newcommand{\cL}{\mathcal L}

\newcommand{\cN}{\mathcal N}

\newcommand{\cS}{\mathcal S}

\newcommand{\cU}{\mathcal U}

\graphicspath{{figs/}}

\newcommand{\norm}[1]{\left \|#1 \right \|}

\newcommand{\uu}{u}
\newcommand{\z}{z}

\newcommand{\x}{x}
\newcommand{\w}{w}

\newcommand{\dd}{d}

\newcommand{\iter}{k}

\newcommand{\n}{n}
\newcommand{\m}{m}

\newcommand{\bz}{\bar{\z}}

\newcommand{\pz}{\z_\perp}

\newcommand{\bw}{\bar{\w}}
\newcommand{\pw}{\w_\perp}

\newcommand{\tpz}{\tilde{\z}_\perp}
\newcommand{\tpw}{\tilde{\w}_\perp}

\newcommand{\lipp}{\beta}

\newcommand{\X}{X}

\newcommand{\f}{\ell}
\newcommand{\phii}{\phi_{i}}

\newcommand{\map}[3]{#1: #2 \rightarrow #3}

\newcommand{\1}{\mathbf{1}}

\newcommand{\agg}{\sigma}

\newcommand{\lagg}{\phi}
\newcommand{\laggi}{\lagg_i}

\graphicspath{{figs/}}

\newcommand{\dyn}{r}
\newcommand{\dyni}{\dyn_i}

\newcommand{\pos}{p}

\newcommand{\ui}{\uu_i}
\newcommand{\uj}{\uu_j}
\newcommand{\dui}{\dot{\uu}_i}
\newcommand{\du}{\dot{\uu}}

\newcommand{\btk}{\tilde{t}^\iter}
\newcommand{\btkp}{\tilde{t}^{\iter+1}}
\newcommand{\tki}{t^{\iter_i}_i}
\newcommand{\tkj}{t^{\iter_j}_j}
\newcommand{\tkpi}{t^{\iter_i+1}_i}

\newcommand{\tr}{\lambda}

\newcommand{\hphii}{\hat{\phi}_i^{\iter}}
\newcommand{\hphij}{\hat{\phi}_j^{\iter}}
\newcommand{\hnfi}{\nabla_2 \hat{\f}_i^{\iter}}
\newcommand{\hnfj}{\nabla_2 \hat{\f}_j^{\iter}}
\newcommand{\hzi}{\hat{\z}_{i}^{\iter}}
\newcommand{\hzj}{\hat{\z}_{j}^{\iter}}
\newcommand{\hwi}{\hat{\w}_{i}^{\iter}}
\newcommand{\hwj}{\hat{\w}_{j}^{\iter}}
\newcommand{\hci}{\hat{\Lambda}_{i}^{\iter}}

\newcommand{\fs}{\f^\sigma}
\newcommand{\fsh}{\f^{\sigma,h}}

\newcommand{\dir}{\dd}
\newcommand{\diri}{\dir_i}

\newcommand{\Gt}{G_2}

\newcommand{\e}{e^\iter}
\newcommand{\ew}{e^{\w,\iter}}
\newcommand{\ez}{e^{\z,\iter}}

\newcommand{\nchi}{n_{\chi}}

\newcommand{\nom}{F}

\newcommand{\ewi}{\ew_i}
\newcommand{\ezi}{\ez_i}
\newcommand{\ei}{\e_i}

\newcommand{\ewj}{\ew_j}
\newcommand{\ezj}{\ez_j}

\newcommand{\pol}{\mathrm{q}}

\newcommand{\poli}{\pol_i}

\newcommand{\cart}{\mathrm{p}}

\newcommand{\carti}{\cart_i}

\newcommand{\pid}{g} 

\newcommand{\cm}{\upsilon}
\newcommand{\vq}{\mathrm{v}}

\newcommand{\np}{S}

\newcommand{\point}{s_\ell}

\newcommand{\nd}{D}

\newcommand{\tdir}{\Delta\dir} 

\newcommand{\circconc}{\text{C}}

\newcommand{\rr}{\mathrm{r}}

\newcommand{\rot}{\mathcal{R}}

\def\algo/{{\scshape Triggered Aggregative Tracking Feedback}}
\def\algoAut/{{Aggregative Tracking Feedback}}

\def\er/{Erd\H{o}s-R\'enyi}

\title{Multi-Robot Target Monitoring and Encirclement\\ via Triggered Distributed Feedback Optimization}

\author{%
	Lorenzo Pichierri, Guido Carnevale, Lorenzo Sforni and Giuseppe Notarstefano
	\thanks{This study was carried out within the MOST - Sustainable Mobility National Research Center and received 
	funding from the European Union Next-GenerationEU (PIANO NAZIONALE DI RIPRESA E RESILIENZA (PNRR) - MISSIONE 4 COMPONENTE 2, INVESTIMENTO 1.4 - D.D. 1033 17/06/2022, CN00000023). This manuscript reflects only the authors' views and opinions, neither the European Union nor the European Commission can be considered responsible for them. 
	}%
	\thanks{Authors are with the Department of Electrical, 
		Electronic and Information Engineering, University of Bologna, Bologna, Italy.
		Email: \texttt{\{name.surname\}@unibo.it}.}%
}

\begin{document}

	\maketitle
	
\begin{abstract}
	We design a distributed feedback optimization strategy, embedded into a modular ROS~2 control architecture, which allows a team of heterogeneous robots to cooperatively monitor and encircle a target while patrolling points of interest. Relying on the aggregative feedback optimization framework, we handle multi-robot dynamics while minimizing a global performance index depending on both microscopic (e.g., the location of single robots) and macroscopic variables (e.g., the spatial distribution of the team). The proposed distributed policy allows the robots to cooperatively address the global problem by employing only local measurements and neighboring data exchanges. These exchanges are performed through an asynchronous communication protocol ruled by locally-verifiable triggering conditions. We formally prove that our strategy steers the robots to a set of configurations representing stationary points of the considered optimization problem. The effectiveness and scalability of the overall strategy are tested via Monte Carlo campaigns of realistic Webots ROS~2 virtual experiments. Finally, the applicability of our solution is shown with real experiments on ground and aerial robots.	
\end{abstract}

\vspace*{-.1cm}
\begin{IEEEkeywords}
	Distributed Robot Systems;
	Multi-Robot Systems;
	Optimization and Optimal Control;
	Distributed Optimization.
\end{IEEEkeywords}

\vspace*{-.1cm}
	
\section{Introduction}
\label{sec:introduction}
Multi-robot systems have gained a lot of interest in the last years for
their capability of addressing complex tasks via
cooperation~\cite{rubio2019review}.
We investigate a framework in which a team of robots wants to
cooperatively monitor and encircle a common target while patrolling
specific points of interest.
\begin{figure}
  \centering
  \includegraphics[width=0.5\columnwidth]{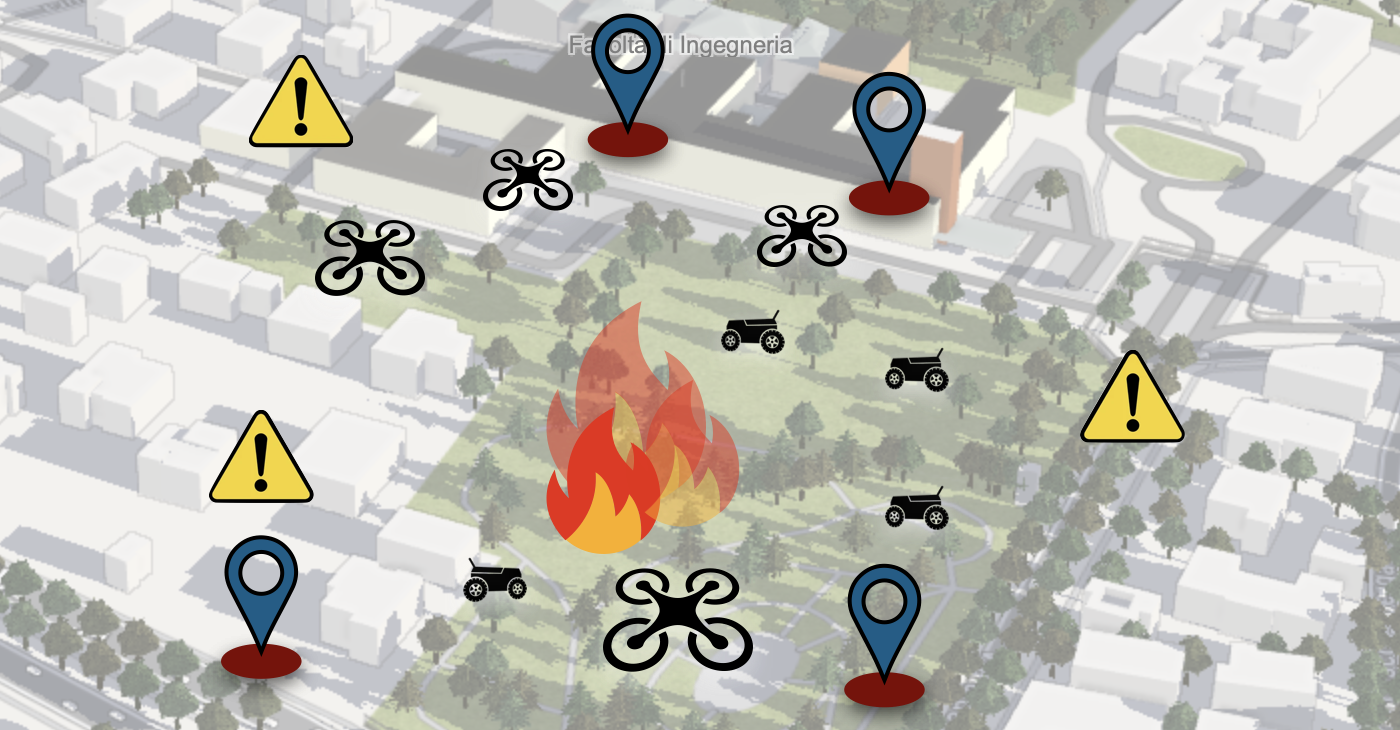}
  \caption{
  A heterogeneous team of aerial and ground robots trying to cooperatively encircle a target while patrolling points of interest.}

  \label{fig:intro}
\end{figure}

\paragraph*{Related Work}

The encirclement problem has been mainly addressed in the
  literature by framing the task as a formation control problem in
  which robots implement a distributed protocol to place themselves
  around the
  target~\cite{guo2010cooperative,franchi2016decentralized,deshpande2018self,ma2020multi,meng2023bi}.
  In these formulations, the team is not able to adapt its
  configuration to multi-objective tasks as, e.g., monitoring sensitive
  spots and taking into account density functions in the environment.
  Another approach has been recently proposed
  in~\cite{sinhmar2023guaranteed}, where target encirclement is
  achieved by leveraging only the sensing capabilities of the robots.
The monitoring and encirclement problems are deeply related to surveillance tasks. 
In this context, a coverage control approach has been used
in~\cite{nigam2011control,adaldo2017cooperative,petrlik2019coverage}.
In~\cite{zhang2010decentralized}, a distributed, multi-objective
algorithm is designed to detect intruders and protect sensitive areas.
Authors in~\cite{raboin2013model,mullins2015adversarial} consider a
team of autonomous vehicles aimed at maximizing the amount of time
needed by an intruder to reach a target.
Cooperative robotics tasks are often solved via distributed optimization frameworks, see, e.g.,~\cite{shorinwa2024distributeda, shorinwa2024distributedb, notarstefano2019distributed}. Our
paper models the multi-robot, multi-objective task by resorting to
the formalism of aggregative optimization.
Indeed, such an emerging framework is gaining increasing interest in
the context of cooperative
robotics~\cite{pichierri2023distributed,testa2023tutorial}.  In this
field, a network of robots aims to minimize an objective function
given by the sum of local functions which depend on both local
decision variables (e.g., the position of the robot) and an
aggregation of all of them (e.g., the barycenter of the team).
The first work studying this new class of problems is~\cite{li2021distributed}, where an unconstrained and static scenario is considered.
In~\cite{li2021distributedOnline,carnevale2022distributed}, the problem has been extended to deal with constrained and time-varying scenarios.
The feedback optimization approach (see the survey~\cite{hauswirth2021optimization}) is studied in the aggregative scenario in~\cite{carnevale2024nonconvex}  and~\cite{carnevale2022aggregative}.
The distinguishing feature of these works lies in their ability to take into  account
the dynamics of local plants whose state coincides with the decision variable of the optimization problem.
Moreover, the gradients of the unknown functions of the optimization problem are assumed to be measured only according to the current configuration.

\paragraph*{Contributions}
In this work, we propose a distributed strategy to solve
monitoring and encirclement tasks for (heterogeneous)
multi-robot systems.
We model the problem by resorting to the formalism of distributed feedback optimization.
More in detail, the optimization problem 
is formulated by leveraging the recently emerged aggregative structure and each
robot dynamics is modeled as a nonlinear control
system.

The main contribution of the paper is twofold. 

\noindent First, we propose a
novel distributed feedback optimization algorithm named \algo/ to
handle concurrently the control of robots and the global optimization
goal.
The proposed strategy extends the distributed feedback optimization
law designed in~\cite{carnevale2024nonconvex} with an event-triggered
communication scheme in which each robot asynchronously sends data to
its neighbors according to a locally verifiable triggering condition.
We formally prove that our strategy steers the network of robots to a steady-state configuration corresponding to a stationary point of an underlying nonconvex aggregative optimization problem.
Further, we also guarantee that Zeno's behavior is excluded.

\noindent Second,
the proposed algorithmic strategy is tailored for cooperative robotics applications.
To this end, (i) we design the aggregative optimization objective function to properly model some specific robotic tasks, i.e., monitoring, encirclement and safe placement,
and (ii) we develop the full-stack control architecture required to deploy the proposed event-triggered distributed feedback optimization strategy on teams of heterogeneous robots.
Such a distributed strategy is embedded into a modular ROS~2 control architecture, 
based on the recently emerged frameworks \textsc{Crazychoir}~\cite{pichierri2023crazychoir} and \textsc{ChoiRbot~\cite{testa2021choirbot}} tailored for distributed robotics applications.

We validate the effectiveness of our distributed control architecture in both virtual and real-world experimental settings.
Specifically, our distributed strategy is tested through realistic
Webots virtual experiments within a ROS~2 framework and it is further
deployed on real hardware, including Crazyflie nano-quadrotors and
TurtleBot ground robots.
Additionally, we conduct Monte Carlo campaigns to study the communication load and highlight the scalability of the proposed solution.
Unlike other works in the literature, our strategy is particularly suited for dynamic environments
thanks to the peculiar structure of the
aggregative framework, which allows for cooperative encirclement without an offline allocation phase.

This work is a significant extension of our conference version~\cite{pichierri2024distributed}. 
The main improvements over~\cite{pichierri2024distributed} are as follows.
\begin{enumerate} 
	\item We replace the unrealistic continuous-time inter-agents communication used in~\cite{pichierri2024distributed} with an asynchronous mechanism governed by locally verifiable triggering conditions. 
	We also provide a formal proof establishing the stability and attractiveness properties of the obtained distributed event-triggered strategy; 
	\item We address a more complex scenario using density functions to model dangerous areas (rather than single points) and a more refined strategy for the encirclement task;
	\item We introduce a suitable mechanism based on Control Barrier Functions (CBFs) to avoid collisions among the robots in a distributed manner;
	\item We test the overall distributed architecture via real experiments and Monte Carlo campaigns of virtual experiments with heterogeneous teams of robots, whereas~\cite{pichierri2024distributed} only presented a simplified simulation on aerial teams.
\end{enumerate}

\paragraph*{Organization} 
The paper unfolds as follows. 
In Section~\ref{sec:preliminaries}, we introduce the considered problem setup.
Section~\ref{sec:algorithm} introduces our distributed strategy and its theoretical properties.
In Section~\ref{sec:modeling}, we detail how to model multi-robot
monitoring and encirclement tasks via aggregative optimization.
Virtual and real experiments are provided in
Section~\ref{sec:simulations} and Section~\ref{sec:experiments},
respectively.

\paragraph*{Notation} 

The symbol $\col (v_1, \ldots, v_n)$ denotes the concatenation of the vectors $v_1, \ldots, v_n$.
Given $N$ scalars $v_1, \dots, v_N$, the symbol $\diag\{v_1,\dots,v_N\}$ denotes the diagonal matrix with the $v_i$ on the $i$-th diagonal component.
$I_m$ is the identity matrix in $\R^{m\times m}$.
$1_N$ and $0_N$ are the vectors of $N$ ones and zeros, respectively, while $\1_{N,d} \coloneqq 1_N \otimes I_d$, where $\otimes$ denotes the Kronecker product.
Dimensions are omitted whenever they are clear from the context.
Given $\ell: \R^{n_1} \times \R^{n_2} \to \R^n$, we define
$\nabla_1 \ell(\x,y) \coloneqq \tfrac{\partial}{\partial s}\ell(s,y)|_{s =
  x}$ and
$\nabla_2 \ell(\x,y) \coloneqq \tfrac{\partial}{\partial s}\ell(\x,s)|_{s =
  y}$.
Let $X \subseteq \R^{n}$ and $x \in \mathbb{R}^n$, then $\|x\|_X \coloneqq \inf_{y \in X}\norm{x - y}$.
Time-dependency is omitted in every dynamical system $\dot{x}(t) = r(x(t))$.

\newpage
\section{Problem Formulation}
	\label{sec:preliminaries}

	In this section, we detail our distributed feedback strategy.
	As it will become clearer in the next, the proposed method relies on information exchanges among the robots of the team. 
	Thus, before introducing the whole setup, we establish the communication model connecting the cooperating robots.

	\subsection{Communication Model}

	We assume that robots can exchange information according to a communication network modeled as a directed graph
	$\cG = (\until{N},\cE)$ 
	in which $\cE \subseteq \until{N}\times \until{N}$ is the edge set.
	The graph $\cG$ models the communication in the sense that there is an edge $(i,j)\in \cE$ if robots $j$ is able
	to send information to robot $i$. 
	For each node $i$, the set of (in)-\emph{neighbors} is denoted by $\cN_i =\{j\in \until{N} \mid (i,j)\in\cE\}$.
	We also associate a so-called weighted adjacency matrix $\cA \in \R^{N \times N}$ matching the graph, i.e., a matrix with $(i,j)$-entry $a_{ij} > 0$ if $(j,i) \in \cE$, otherwise $a_{ij} = 0$. 
	The weighted in-degree and out-degree of agent $i$ are
	defined as $deg_i^{\text{in}} = \sum_{j\in \cN_i}a_{ij}$ and
	$deg_i^{\text{out}}= \sum_{j\in \cN_i}a_{ji}$, respectively. 
	Finally, we associate to $\cG$ the so-called
	Laplacian matrix defined as $\cL\coloneqq\mathcal{D}^{\text{in}} - \cA$, where
	$\mathcal{D}^{\text{in}} \coloneqq \text{diag}(deg_1^{\text{in}},\ldots,deg_N^{\text{in}}) \in \R^{N \times N}$.
	The next assumption formalizes the class of graphs considered in this work.
	\begin{assumption}\label{ass:network}
		$\cG$ is strongly connected and weight-balanced, namely $deg_i^{\text{in}}  =  deg_i^{\text{out}}$ for all $i  \in  \until{N}$.\oprocend
	\end{assumption}

	\subsection{Multi-Robot Model}
	\label{sec:dynamics}

	The dynamics of each robot $i$ reads as 
	\begin{align}
		\dot{\x}_i = \dyni(\x_i,\ui),\label{eq:local_plant}
	\end{align}
	where $\x_i \in \R^{n_i}$ and $\ui \in \R^{m_i}$ are the state and the input of robot $i$, respectively, and $\dyni: \R^{n_i} \times \R^{m_i} \to \R^{n_i}$ represents its dynamics.
	As customary in the field of feedback optimization, the dynamics described in~\eqref{eq:local_plant} is already stabilized, i.e., robot $i$ is endowed with a local low-level controller embedded into the map $\dyni$.
    As formalized in the next assumption, this controller ensures exponential stability properties for dynamics~\eqref{eq:local_plant}.
	\begin{assumption}\label{ass:steady_state}
		For all $i \in \until{N}$, there exist $h_i: \R^{m_i} \to \R^{n_i}$ and $\delta_1, \delta_2 > 0$ such that, for all $\bar{\uu}_i \in \R^{m_i}$, it holds $0 = r_i(h_i(\bar{\uu}_i),\bar{\uu}_i)$ and the trajectories of~\eqref{eq:local_plant} satisfy
		\begin{align*}
			\norm{\x_i(t) - h_i(\bar{\uu}_i)} \leq \delta_1\norm{\x_i(0) - h_i(\bar{\uu}_i)}\exp(-\delta_2t),
		\end{align*}
		for all $\x_i(0) \in \R^{n_i}$ and $t \ge 0$.		
		Further, there exist $\lipp_h, \lipp_\dyn > 0$ such that
		\begin{align*}
		\|h_i(\ui)-h_i(\ui^\prime)\| &\le \lipp_h \|\ui-\ui^\prime\|
		\\
		\|r_i(\x_i,\ui)-r_i(\x_i^\prime,\ui^\prime)\| &\le \lipp_\dyn \|\x_i - \x_i^\prime\|
		+ \lipp_\dyn \|\ui-\ui^\prime\|,
		\end{align*}
		for all $\x_i,\x_i^\prime \in \mathbb{R}^{n_i}$, $\ui,\ui^\prime \in \mathbb{R}^{m_i}$, and $i \in \until{N}$. \oprocend
	\end{assumption}
	The next section formalizes the underlying aggregative optimization problem that we use to model the cooperative monitoring and encirclement task.

	\subsection{Modelling Multi-Robot Target Monitoring and Encirclement as Aggregative Feedback Optimization}
	\label{sec:opt_problem}

\begin{figure}
		\centering
		\includegraphics[width=0.5\columnwidth]{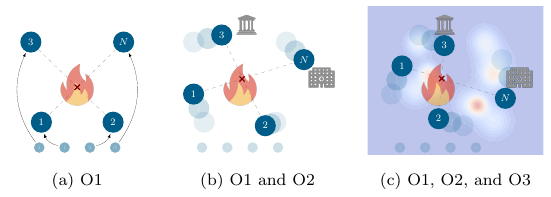}
		\vspace*{-0.2cm}
		\caption{An illustrative example of desired
                    placement. In the three subfigures, the final
                    configuration is shown when considering only
                    objective \textit{O1} (a), objectives \textit{O1} and \textit{O2} (b), and
                    all the three, \textit{O1}, \textit{O2}, and \textit{O3} (c).}
		\label{fig:example_2}
	\end{figure}

	To address the monitoring and encirclement problem, each robot $i \in \until{N}$ should steer its state $\x_i$ toward configurations that result from a tradeoff between potentially conflicting global and local objectives.
	The desired behavior of the team can be outlined by the following objectives:
	\begin{enumerate}[\it O1)]
		\item \textit{(Global)} Cooperative target encirclement: steering the team in a configuration enclosing a given target;
		
		\item \textit{(Local)} Multi-spot monitoring: placing itself as close as possible to a specific point of interest (e.g., a sensitive structure during a wildfire);
		
		\item \textit{(Local)} Safe placement: moving toward a safe configuration with respect to a given potential modeling the (possible) presence of dangers (due to, e.g., the high temperature generated by a wildfire).
	\end{enumerate}
	An illustrative example of the optimal configuration we want to achieve is provided in Fig.~\ref{fig:example_2}.
	To address this problem, we resort to the formalism of the so-called aggregative optimization framework.
	In this setup, the robots aim to solve
	\begin{align}
			\min_{x \in \R^{n}} \: & \: \sum_{i =1}^N \f_i(x_i,\agg(x)),
		\label{eq:aggregative_problem_intermediate}
	\end{align}
	where $x = \col(x_1,\ldots,x_N) \in \R^{n}$ with $n \coloneqq \sum_{i=1}^N n_i$, and each cost $\ell_i: \R^{n_i} \times \R^{d} \to \R$ depends on both the local decision variable of robot $i$ and on an \emph{aggregative variable} $\sigma: \R^{\n} \to \R^{d}$ given by
	\begin{align*}
		\agg(x) \coloneqq \dfrac{1}{N}\sum_{i=1}^N \laggi(x_i),
	\end{align*}
	with $\laggi(x_i):\R^{n_i} \to \R^d$ for each $i\in\until{N}$.
	The aggregative variable $\sigma(x)$ will be suitably defined to properly capture objective \textit{O1}.
	To steer the robots towards a configuration solution of problem~\eqref{eq:aggregative_problem_intermediate} while also considering the robots' dynamics, we resort to the concept of feedback optimization. 
	Thus, we focus on a suitably modified version of problem~\eqref{eq:aggregative_problem_intermediate} with the steady-state maps $h_i$ (see Assumption~\ref{ass:steady_state}), namely 
	\begin{align}
		\begin{split}
			\min_{x \in \R^{n}, u \in \R^{m}} \: & \: \sum_{i =1}^N \f_i(x_i,\agg(x)).
			 \\
		  	\subj \: & \: x_i = h_i(u_i), \forall i \in \until{N},
		\end{split}
		\label{eq:aggregative_problem}
	\end{align}
	where $u = \col(u_1,\ldots,u_N) \in \R^{m}$ with $m\coloneqq\sum_{i=1}^N m_i$.
	We also introduce the overall objective function $\fs: \R^{n} \to \R$ and its reduced version $\fsh: \R^{m} \to \R$ defined as
	\begin{align*}
		\fs(x) &\coloneqq \sum_{i =1}^N \f_i(x_i,\agg(x))
		\\
		\fsh(u) &\coloneqq \sum_{i =1}^N \f_i(h_i(u_i),\agg(h(u))).
	\end{align*}
	The next assumption characterizes the class of problems investigated in this paper.
	\begin{assumption}\label{ass:lipschitz}
		The function $\fsh$ is radially unbounded, differentiable and its gradient is Lipschitz continuous with constant $\lipp_0 > 0$.
		Moreover, the gradient of $\fs$ is Lipschitz continuous with constant $\lipp_0^\agg$.
		Furthermore, for all $i \in \until{N}$, $\nabla_1 \f_i$, $\nabla_2 \f_i$, and $\phii$ are Lipschitz continuous with constants $\lipp_1,\lipp_2, \lipp_3 > 0$, respectively.  \oprocend
	\end{assumption}
	Throughout this paper, we consider a cooperative, distributed scenario in which robots have limited knowledge of problem~\eqref{eq:aggregative_problem}. 
	Thus, in order to solve the optimization problem, they have to iteratively exchange suitable information via a communication network.
	Moreover, as customary in feedback optimization, the analytic expression of the local costs $\f_i$ and aggregation rules $\phii$ are not available to the robots, they can be only measured according to the current states $\x_i$.
	Specifically, each robot $i$ can only access $\nabla_1 \f_i (\x_i,\hat{\sigma}_i)$, $\nabla_2 \f_i(\x_i,\hat{\sigma}_i)$, $\phii(\x_i)$, $\nabla\phii (\x_i)$, and $\nabla h_i(\uu_i)$ where $\x_i$ and $\ui$ are its current state and input variables, respectively, while $\hat{\sigma}_i \in \R^d$ is its local estimate of $\sigma(\x)$.

	\section{\algo/: Distributed Algorithm Description}
	\label{sec:algorithm}

	In order to steer the robots modeled as~\eqref{eq:local_plant} toward a configuration representing a solution of problem~\eqref{eq:aggregative_problem}, we propose a distributed feedback optimization strategy based on~\cite{carnevale2024nonconvex} and incorporating an event-triggered communication protocol inspired by~\cite{carnevale2023triggered}.
	In the following, we first recall the design idea behind the distributed feedback optimization law proposed in~\cite{carnevale2024nonconvex} and, then, we describe our event-triggered communication protocol.
	The section concludes providing the stability and convergence properties of the obtained networked closed-loop system.

	\subsection{Continuous-time communication: \algoAut/}
	\label{sec:algo_aut}

	This section recalls the distributed feedback optimization law proposed in~\cite{carnevale2024nonconvex}.
	The strategy is derived by considering the so-called reduced version of problem~\eqref{eq:aggregative_problem} which is obtained by considering each robot $i$ in the steady-state configuration $\x_i = h_i(\ui)$ (cf. Assumption~\ref{ass:steady_state}).
	Such a problem reads as
	\begin{align}\label{eq:tilde_aggregative_problem}
	\min_{u \in \R^m} \sum_{i=1}^N \f_i(h_i(u_i),\sigma(h(u))).
	\end{align}
	Then, by deriving $\sum_{i=1}^N \f_i(h_i(u_i),\sigma(h(u)))$ with respect to $\ui$, each robot $i$ may implement the gradient-flow controller
	\begin{align}\label{eq:centralized_aggregative_with_u}
	\dui &= 
	 - \nabla h_i(\ui) \Bigg(\nabla_1 \f_i(h_i(\ui),\sigma(h(\uu))) 
	+\dfrac{\nabla \phii(h_i(\ui))}{N} \sum_{j=1}^{N}\nabla_2 \f_j(h_j(\uj),\sigma(h(\uu))) 
	\Bigg),
	\end{align}
	First, keeping in mind the desired distributed paradigm of our strategy, the global quantities $\sigma(\x)$ and $\sum_{j=1}^{N}\nabla_{2}\f_j(\x_j,\sigma(x))$ are not available to robot $i$.
	Second, as customary in feedback optimization, robot $i$ can only measure $\nabla_1 \f_i(\x_i,\hat{\sigma}_i)$, $\nabla \phii(\x_i)$, and $\sum_{j=1}^{N}\nabla_2 \f_j(\x_j,\hat{\sigma}_j)$, where $\hat{\sigma}_i \in \R^{d}$ is the local estimate of agent $i$ about $\sigma(\x)$.
	Hence, dynamics~\eqref{eq:centralized_aggregative_with_u} is modified by using the available measures and two auxiliary variables $\w_i, \z_i \in \R^{d}$ compensating the lack of knowledge about $\sigma(\x)$ and $\frac{1}{N}\sum_{j=1}^{N}\nabla_2 \f_j(\x_j,\hat{\sigma}_j)$ via a continuous-time tracking mechanism.
	The overall strategy thus becomes
	\begin{subequations}\label{eq:algo_aut}
		\begin{align}
			\dui  &= -\alpha_1\diri(\x_i,\ui,\w_i,\z_i)
			\label{eq:x_i_dynamics_aut}
			\\
			\dot{\w}_i &= -\dfrac{1}{\alpha_2}\sum_{j\in\cN_i}a_{ij}\left(\w_i + \phii(\x_i) - \w_j - \phi_j(\x_j)\right)
			\label{eq:w_i_dynamics_aut}
			\\
			\dot{\z}_i &=
			-\dfrac{1}{\alpha_2}\sum_{j\in\cN_i}  a_{ij}(z_i  +  \nabla_{2}\f_i(\x_i,\w_i+ \phii(\x_i)) - z_j - \nabla_{2}\f_j(\x_j,\w_j+ \phi_j(\x_j)))
			  \label{eq:z_i_dynamics_aut},
		\end{align}
	\end{subequations}
	where $\diri: \R^{n_i} \times \R^{m_i} \times \R^{d} \times \R^{d} \to \R^{m_i}$ reads as
	\begin{align*}
		\diri(x_i,u_i,w_i,z_i) 
		\coloneqq 
		\nabla h_i(u_i)
		\left(
			\nabla_1 \f_i(x_i,w_i+\phii(x_i))
			+
			\nabla\phii(x_i)(\nabla_2\f_i(x_i,w_i+\phii(x_i)) + z_i)
		\right),
	\end{align*}
	for all $i \in \until{N}$.
	With the concepts of singular perturbations theory in mind (see, e.g.,~\cite[Ch.~9]{khalil2002nonlinear}), we add the parameters $\alpha_1, \alpha_2 >0$ to adjust the speed of $\ui$ and $(\w_i,\z_i)$, respectively. Specifically, they are designed to make the optimization-oriented part~\eqref{eq:x_i_dynamics_aut} and the consensus-oriented one~\eqref{eq:w_i_dynamics_aut}-\eqref{eq:z_i_dynamics_aut} respectively slower and faster than the robot dynamics~\eqref{eq:local_plant}.
	In order to implement dynamics~\eqref{eq:algo_aut}, each robot $i$ needs to exchange two vectors (namely, $w_i + \phii(\x_i)$ and $z_i + \nabla_2\f_i(x_i,w_i + \phii(x_i))$) in $\R^d$ only with its neighbors.
	Thus, the proposed solution is purely distributed and requires exchanges
	of $2d$ floats.
	However, these exchanges require a continuous-time communication among the robots.
	Clearly, since real robots require time-slotted communication, this prevents its practical implementation and, thus, motivates the event-triggered protocol described in the next section.

	\subsection{\algo/}
	\label{sec:event_triggered}

	Since a continuous communication protocol cannot be practically implemented while a synchronous one may give rise to a non-efficient usage of the robot
	resources, we opt for an approach including an event-triggered communication protocol.
	This choice allows robots to exchange information only when really needed and, thus, to reduce the communication burden.
	Let $\{\tki\}_{k_i \in \N}$, be the sequence of time instants at which agent $i$ sends the local quantities $(\w_i + \phii(\x_i),\z_i + \nabla_2 \f_i(\x_i,\w_i + \phii(\x_i)))$ to its neighbors $j\in\cN_i$. 
	Consistently, at time $\tkj$, agent $i$ receives the updated variables from its neighbor $j \in \cN_i$.
	Let $\{\btk\}_{k\ge 0}$ be the ordered sequence of all the triggering times occurred in the network. 
	Then, we introduce
	\begin{subequations}\label{eq:hat}
	\begin{align}
		\hwi
		&
		\coloneqq  
		\w_i (t) \Big|_{
			t= \sup\limits_{\iter_i \in \N} \left\{ \tki \le \btk\right\}
			}, 
		\hspace{.05cm} 
		\hzi
		\coloneqq  
		\z_i (t) \Big|_{
				t= \sup\limits_{\iter_i \in \N} \left\{ \tki \le \btk\right\}
				}
		\\
		\hphii
		& 
		\coloneqq \phii(\x_i(t)) \Big|_{
			t= \sup \limits_{\iter_i \in \N}\left\{ \tki \le \btk\right\}
			}
		\\
		\hnfi
		& 
		\coloneqq \nabla_2 \f_i(\x_i(t),\w_i(t)+\phii(\x_i(t))) \Big|_{
			t= \sup \limits_{\iter_i \in \N}\left\{ \tki \le \btk\right\}
		},
		\end{align}
	\end{subequations}
	for all $i \in\until{N}$.
	The above quantities represent the most updated values in the network within $[\btk,\btkp)$.
	We also introduce the errors $\ewi,\ezi \in\R^{d}$ and $\ei \in\R^{2d}$ defined as
	\begin{align}
		\ei 
		&\coloneqq
		\begin{bmatrix}
			\ewi
			\\
			\ezi
		\end{bmatrix} 
		 \coloneqq 
		\begin{bmatrix} 
			\w_i - \hwi + \phii(\x_i) - \hphii
			\\
			\z_i  -  \hzi
			 +  \nabla_2 \f_i(\x_i,\w_i  +  \phii(\x_i)) 	 -  \hnfi
		\end{bmatrix}.
		\label{eq:e_i}
	\end{align}
	We are ready to present our distributed feedback optimization law termed \algo/, giving rise to the closed-loop system resumed in Algorithm~\ref{alg:closed_loop} from the perspective of agent $i$.
	\begin{algorithm}
		\begin{algorithmic}
		\State \textbf{Initialization}:
		\State 
		$\x_i(0) \in \R^{\n_i}, 
		\ui(0) \in \R^{\m_i}, 
		\w_i(0) =\hat{\w}^0_i = 0, 
		\z_i(0) = \hat{\z}^0_i = 0$, 
		$\hat{\phi}_i^0 = \phii(\x_i(0))$, 
		$\nabla_2\hat{\f}_i^0 = \nabla_2\f(\x_i(0),\w_i(0) + \phii(\x_i(0)))$, $\xi_i(0) \ne 0$.
		\vspace{2mm}

		\State \textbf{Evolution}:
		\begin{subequations}\label{eq:local_closed_loop}
			\begin{align}
				\dot{\x}_i &= \dyni(\x_i,\ui)
				\\
				\dui  &= - \alpha_1\diri(\x_i,\ui,\w_i,\z_i)
				\\
				\dot{\w}_i &= -\dfrac{1}{\alpha_2}\sum_{j\in\cN_i}a_{ij}\left(\hwi + \hphii - \hwj - \hphij\right)
				\label{eq:local_closed_loop_w_i}
				\\
				\dot{\z}_i &=
				-\dfrac{1}{\alpha_2}\sum_{j\in\cN_i}  a_{ij}(\hzi  +  \hnfi- \hzj - \hnfj)
				\label{eq:local_closed_loop_z_i}
				\\
				\dot{\xi}_i &= -\nu \xi_i.\label{eq:xi}
			\end{align}
		\end{subequations}
		\State \textbf{Triggering Mechanism}:
		\vspace{2mm}

		\If{$\norm{\ei} > \tr \norm{\diri(\cdot)} 
		+ |\xi_i|$, with $\ei$ from \eqref{eq:e_i}}  
			\State $\hwi = w_i$, $\hphii = \phii(x_i)$ 
			\State $\hzi = z_i$, $\hnfi = \nabla_2\f_i(\x_i,\w_i + \phii(x_i))$
			\vspace{2mm}

			\State send $(\hwi + \hphii,\hzi + \hnfi)$ to neighbors
		\EndIf
		\end{algorithmic}
		\caption{Closed-Loop System - Robot $i$}
		\label{alg:closed_loop}
	\end{algorithm}

	In particular, in order to perform communication only when needed, each robot $i$ chooses the next triggering time instant $\tkpi$ according to the locally verifiable condition
	\begin{align}\label{eq:triggering_law}
		\tkpi \coloneqq \inf_{t > \tki} \Big \{
			\norm{\ei(t)} &> \tr \norm{\diri(\x_i(t),\uu_i(t),\w_i(t),\z_i(t))} 
			+ |\xi_i(t)| \Big \},
	\end{align}
	where $\tr > 0$ is a tuning parameter determining the communication frequency, while $\xi_i \in \R$ is a local, auxiliary variable having dynamics~\eqref{eq:xi} where $\nu > 0$ rules its decay.
	The rationale for the triggering mechanism is to \emph{(i)} keep the triggered scheme  as close as possible to the not implementable dynamics~\eqref{eq:algo_aut}, and \emph{(ii)} avoid the so-called Zeno behavior (see, e.g.,~\cite{goebel2009hybrid})
	As formally shown next, if each $\xi_i$ is initialized to a nonzero value, the arising closed-loop system with triggering law~\eqref{eq:triggering_law} does not incur in the Zeno behavior.
	As one may expect, we note that, differently from the strategy in~\eqref{eq:algo_aut}, now robot $i$, in the consensus-oriented part, uses the sampled quantities $\hwj + \hphij$ and $\hzj + \hnfj$ of the neighbors $j \in \cN_i$ instead of their current counterparts $\w_j + \phi_j(\x_j)$ and $\z_j + \nabla_2 \f_j(\x_j,\w_j + \phi_j(\x_j))$.
	Notice that robot $i$ does not use its own quantities $\w_i + \phii(\x_i)$ and $\z_i + \nabla_2 \f_i(\x_i,\w_i + \phii(\x_i))$ in the consensus mixing terms, although they are available. It rather uses their sampled version to preserve the mean conservation property of dynamics~\eqref{eq:local_closed_loop_w_i}--\eqref{eq:local_closed_loop_z_i}.

\begin{figure}
		\centering
		\includegraphics[width=0.5\columnwidth]{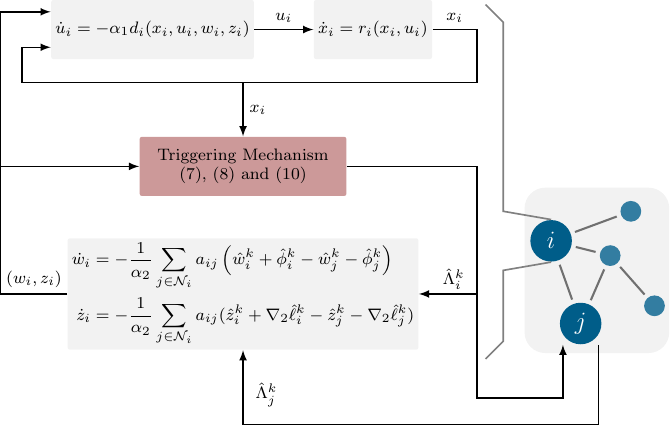}
		\caption{Block diagram representation of Algorithm~\ref{alg:closed_loop}, where the exchanged data are denoted by $\hci \coloneqq (\hwi + \hphii, \hzi + \hnfi)$.}
		\label{fig:algo_scheme}
	\end{figure}

	Finally, the local dynamics in Algorithm~\ref{alg:closed_loop} is graphically depicted in Fig.~\ref{fig:algo_scheme} in terms of block diagram.
	\subsection{Stability and Convergence Analysis}
	The next theorem ensures the convergence and stability properties of the overall closed-loop system arising from the local dynamics in Algorithm~\ref{alg:closed_loop}.
	To state these properties, let $X \subset \R^{n} \times \R^{m}$ be the set of stationary points of problem~\eqref{eq:aggregative_problem}, and let us define
	\begin{align*}%
		\w  &\coloneqq  \begin{bmatrix}
			\w_1\\
			\vdots
			\\
			\w_N
	\end{bmatrix}, \hspace{.1cm} \z  \coloneqq  \begin{bmatrix}
			\z_1\\
			\vdots\\
			\z_N 
		\end{bmatrix}, \hspace{.1cm} \xi \coloneqq \begin{bmatrix}
			\xi_1 
			\\
			\vdots 
			\\
			\xi_N
		\end{bmatrix}
		\\
		\pi^\w(x)  &\coloneqq  \begin{bmatrix}
			-\phi_1(\x_1) + \sigma(\x)
			\\
			\vdots 
			\\
			-\phi_N(\x_N) + \sigma(\x)
		\end{bmatrix}
		\\
		\pi^z(x)  &\coloneqq  \begin{bmatrix}
			-\nabla_2 \f_1(x_1,\sigma(x)) + \dfrac{1}{N}\sum_{j=1}^{N}\nabla_2 \f_j(x_j,\sigma(\x)) 
			\\
			\vdots
			\\
			-\nabla_2 \f_N(x_N,\sigma(x)) + \dfrac{1}{N}\sum_{j=1}^{N}\nabla_2 \f_j(x_j,\sigma(x)) 
		\end{bmatrix}
		\\
		\\
		\mathcal{S}  &\coloneqq  \{\col(\x,\uu,\w,\z,\xi) \in \R^{n+m+N(2d+1)}
		\mid  \1\T \w  =  0, \1\T \z  =  0, \xi \neq 0 \}
		\\
		\mathcal{X} &\coloneqq \{\col(\x,\uu,\w,\z,\xi) \in \R^{n+m+N(2d+1)}
		\mid (\x,\uu) \in \X, \w = \pi^\w(\x), \z = \pi^\z(\x),\xi = 0\}.
	\end{align*}
	\begin{theorem}\label{th:convergence}
		Consider the closed-loop system arising from~\eqref{eq:local_closed_loop} and let Assumptions~\ref{ass:network},~\ref{ass:steady_state}, and~\ref{ass:lipschitz} hold.
		Then, there exist $\bar{\alpha}_1, \bar{\alpha}_2, \bar{\tr}, \bar{\nu} > 0$ such that, for any $\col(\x_i(0),\uu_i(0),\w_i(0),\z_i(0),\xi(0)) \in \R^{n_i + m_i + 2d + 1}$ such that $\z_i = \w_i = 0$ and $\xi_i(0) \neq 0$ for all $i \in \until{N}$, $\alpha_1 \in (0,\bar{\alpha}_1)$, $\alpha_2\in(0,\bar{\alpha}_2)$, $\tr \in (0,\bar{\tr})$, and $\nu > \bar{\nu}$, the trajectories $(\x(t),\uu(t),\w(t),\z(t))$ are bounded and satisfy
	\begin{align}\label{eq:limit_statement}
		\lim_{t\to\infty} \norm{\begin{bmatrix}
			\x(t)
			\\
			\uu(t)
			\\
			\w(t)
			\\
			\z(t)
			\\
			\xi(t)
		\end{bmatrix}}_{\mathcal{X}}&= 0.
	\end{align}
	Further, given any $\bar{u} \in \R^{m}$ being an isolated stationary point and a local minimum of $\fsh$, the point $\col(h(\bar{u}),\bar{u}, \pi^{\w}(h(\bar{u})),\pi^{\z}(\bar{u}),0)$ is locally asymptotically stable for the closed-loop system resulting from~\eqref{eq:local_closed_loop} restricted to $\cS$. 
	Finally, the Zeno behavior is excluded for each robot $i$.\oprocend
	\end{theorem}
	The proof is provided in Appendix~\ref{sec:proof}.

	\section{Distributed Strategy for Target Monitoring and Encirclement}
	\label{sec:modeling}
	
	In this section, we detail the distributed feedback optimization strategy we proposed in the context of target monitoring and encirclement.
	As already presented in Section~\ref{sec:opt_problem}, the goal of the distributed strategy is to simultaneously achieve a threefold objective,
	namely, $(i)$ encircle a principal target (\textit{O1}), $(ii)$ monitor points of interest (\textit{O2}), and $(iii)$ avoid dangerous areas (\textit{O3}).
	To this end, we start by defining the cost functions of problem~\eqref{eq:aggregative_problem}, which we leverage to model these tasks. 
	Then, we introduce the local dynamics~\eqref{eq:local_plant} of the heterogeneous fleet members, namely $N_a$ aerial and $N_g$ ground robots, and the overall distributed control architecture.

	\subsection{Cost Function Definition}

	We consider an instance of problem~\eqref{eq:aggregative_problem} in which we model the local costs $\f_i(x_i,\agg(x))$ as the sum of three terms, namely
	\begin{align}
		\f_i(x_i,\agg(x))  =  \gamma_1 \f_{i}^{\text{coop}}(\agg(x)) 
								 +  \gamma_2 \f_{i}^{\text{spot}}(x_i)
								 +  \gamma_3 \f_i^{\text{danger}}(x_i),
	\label{eq:cost_fcn}
	\end{align}
	where
	\begin{enumerate}
		\item 	$\f_{i}^{\text{coop}}: \R^{a} \to \R$ models objective 
		\textit{O1}, i.e., the cooperative target encirclement;
		\item $\f_{i}^{\text{spot}}: \R^{n_i} \to \R$ models objective 
		\textit{O2}, i.e., the multi-spots monitoring;
		\item $\f_i^{\text{danger}}: \R^{n_i} \to \R$ models objective 
		\textit{O3}, i.e., placing robots in a safe configuration with respect to an environmental density function.
	\end{enumerate}
	The constants $\gamma_1, \gamma_2, \gamma_3 > 0$ are tuning parameters to prioritize the single tasks.

	In order to define these cost functions, we need to introduce $\carti(\x_i) \in \R^{2}$ as the cartesian planar position of robot $i$ and $\poli(\x_i) \in \R^2$ as the polar coordinates with respect to the target $b \coloneqq \col(b_1,b_2) \in \R^2$, namely 
	\begin{align*}
		\carti(\x_i)& \coloneqq\begin{bmatrix}
			\cart_{1,i}(\x_i)
			\\
			\cart_{2,i}(\x_i)
		\end{bmatrix}
		,\quad 
		\poli(\x_i) \coloneqq \begin{bmatrix}
			\rho_i(x_i)
			\\
			\theta_i(x_i) 
		\end{bmatrix}.
	\end{align*}
	In the following, we detail the modeling of the cost terms $\f_{i}^{\text{coop}}(\agg(x))$, $\f_{i}^{\text{spot}}(x_i)$, and
	$\f_i^{\text{danger}}(x_i)$.

	\subsubsection{Cooperative Target Encirclement}
	\label{subsec:encirclement}

	The cooperative target encirclement is forced by maximizing the so-called circular variance of the robots around target $b$. 
	More in detail, see~\cite{mardia2009directional}, the circular variance $\text{VAR}: \R^{n} \to [0,1]$ is defined as
	\begin{align*}
		\text{VAR}(x) \coloneqq 1 - \circconc(x),
	\end{align*}
	where $\circconc: \R^{n} \to [0,1]$ is the so-called circular concentration and reads as
	\begin{align*}
		\circconc(x)
		\coloneqq \sqrt{
		\left(\dfrac{1}{N}\sum_{i=1}^{N}\cos(\theta_i(x_i))\right)^2 + 
		\left(\dfrac{1}{N}\sum_{i=1}^{N}\sin(\theta_i(x_i))\right)^2}.
	\end{align*}
	Thus, we design the \textit{aggregative variable} $\agg: \R^{n} \to \R^2$ to minimize the circular concentration, namely
	\begin{align*}
		\agg(x) \coloneqq 
		\begin{bmatrix}
			\frac{1}{N}\sum\limits_{i=1}^{N}\cos(\theta_i(x_i))
			\\
			\frac{1}{N}\sum\limits_{i=1}^{N}\sin(\theta_i(x_i))
		\end{bmatrix}.
	\end{align*}
	With this notation at hand, for all robots $i \in\until{N}$, we define the cost function component $\f_{i}^{\text{coop}}(\agg(x))$ as
	\begin{align*}
		\f_{i}^{\text{coop}}(\agg(x)) \coloneqq \norm{\agg(x)}^2.
	\end{align*}

	\subsubsection{Sensitive Spots Monitoring}

	We now design the cost term $\f_{i}^{\text{spot}}(x_i)$ to embed in the strategy the monitoring of some specific locations.
	More in detail, given $\np \in \N$ points of interest, we force each robot $i$ to stay as close as possible to a point $\point$ with $\ell \in \until{\np}$ by defining $\f_{i}^{\text{spot}}(x_i)$ as
	\begin{align*}
		\f_{i}^{\text{spot}}(x_i) = \norm{\carti(x_i) - \point}^2,
	\end{align*}
	where 
	$\point \in \R^2$ is the location of the $\ell$-th point of interest.

	\subsubsection{Safe-Positioning}

	To place robots in a safe location, we introduce a potential $U_i: \R^{n_i} \to \R$ penalizing the configurations close to dangerous locations with an higher costs, namely
	\begin{align*}
		\f_i^{\text{danger}}(x_i) = U_i(x_i),
	\end{align*}
	for all $i \in \until{N}$. %

	\subsection{Heterogeneous Aerial and Ground Mobile Robot Control System}

	Next, we detail the overall system architecture, composed by
	$N_a$ quadrotors and $N_g$ differential-drive ground robots. 
	\subsubsection{Quadrotor: Model and Low-level Control}
	
	The open-loop model of each quadrotor $i \in \until{N_a}$ is modeled as
	\begin{subequations}
	 \label{eq:dyn_crazyflie}
	 \begin{align} 
	  \dot{p}_i &= \vq_i
	  \\
	  \label{eq:dyn_crazyflie:acceleration}
	  \dot{\vq}_i &= g z_{W} - \frac{\cm_{1,i}}{m}\rot(\eta_i)z_{W} - \frac{1}{m} \rot(\eta_i) A \rot(\eta_i)^\top \vq_i 
	  \\
	  \dot{\eta}_i &= \begin{bmatrix}
	   \cm_{2,i}%
	   \\
	   \cm_{3,i}%
	   \\
	   \cm_{4,i}%
	  \end{bmatrix},
	 \end{align}
	\end{subequations}
	where $\pos_i \in \R^3$ is the position of the quadrotor in the world frame
	$\mathcal{W}=\{x_W, y_W, z_W\}$, $\vq_i \in \R^3$ is its linear velocity, while the vector $\eta_i \in \R^3$
	collects the roll, pitch, and yaw angles denoted by $\varphi_i,\vartheta_i, \psi_i$.
	Hence, the whole quadrotor state is $x_i \coloneqq\begin{bmatrix}
	 p_i\T& \vq_i\T& \eta_i\T
	\end{bmatrix}\T \in \R^{9}$ for all $i \in \until{N}$.
	In~\eqref{eq:dyn_crazyflie}, $m\in\R$ is the mass (in kg), and $g$ is the gravity coefficient.
	The rotation matrix is denoted by $\rot(\eta_i) \in SO(3)$.
	The symbol $\cm_i = [\cm_{1,i} \; \cm_{2,i} \; \cm_{3,i} \; \cm_{4,i}]^\top\in\R^4$ denotes the control input of the low-level regulator of the quadrotor. 
	In particular, $\cm_{1,i} \in \R$ denotes the thrust and $\cm_{2,i},\cm_{3,i},\cm_{4,i} \in \R$ the angular rates (according to the $Z-Y-X$ extrinsic Euler representation).
	The control input $\cm_{i}$ is computed via
	\begin{align}
		\cm_i = \pid(\x_i,\uu_i),\label{eq:pid}
	\end{align}
	where  $\pid: \R^{9} \times \R^2 \to \R^4$ is a nonlinear geometric controller fed by the state $\x_i$ and the desired set point $\uu_i \in \R^2$.
	It is possible to show the stabilizing properties of the resulting closed-loop system~\cite{mellinger2011minimum}.
	Thus, in this scenario, the steady-state map $h_i: \R^2 \to \R^9$ reads as
	\begin{align}
	 h_i(\uu_i) = \begin{bmatrix}
	  \uu_i\T  
	  &
	  z_i^{\text{static}}
	  &
	  0_3\T
	  &
	  0_3\T 
	 \end{bmatrix}\T,
	\end{align}
	for all $i \in \until{N_a}$, where $z_i^{\text{static}} \in \R$ is the (static) desired height set-point of quadrotor $i$.
	Therefore, the generic robot dynamics~\eqref{eq:local_plant} is recovered by setting the robot model $\dyni$ as the composition between~\eqref{eq:dyn_crazyflie} and~\eqref{eq:pid}.

	\subsubsection{Ground Mobile Robot: Model and Low-level Control}
	Each differential-drive ground robot $i=1,\ldots, N_g$ follows the unicycle dynamics
	\begin{subequations}
		\label{eq:turlebot}
			\begin{align}
				\dot{p}_{1,i} &= \cm_{1,i} \sin\theta_i
				\\
				\dot{p}_{2,i} &= \cm_{1,i} \cos\theta_i
				\\
				\dot{\psi}_i &= \cm_{2,i}
			\end{align}
	\end{subequations}
	where $p_i = [{p}_{1,i}, {p}_{2,i}]\T \in \R^2$ is the planar position of the ground robot in the world frame and $\psi_i\in\R$ is the heading angle. 
	The symbol $\cm_i = [\cm_{1,i}, \cm_{2,i}]\T \in \R^2$ denotes the stack of the control input, namely, linear velocity $\cm_{1,i}$ and angular velocity $\cm_{2,i}$, respectively.
	These commands are generated via the pose-following kinematic control law applicable to unicycle-type robots, proposed in~\cite{park2011smooth}, fed by both the desired set point $\uu_i \in \R^2$, provided by our strategy, and a static reference for the heading angle.
	This pose-following controller renders the desired pose an asymptotically stable equilibrium for the closed-loop system. Hence, for all $i \in \until{N_g}$, the steady-state map $\map{h_i}{\R^2}{\R^3}$ in Assumption~\ref{ass:steady_state} is
	\begin{align*}
		h_i(\uu_i) = \begin{bmatrix}
		 \uu_i\T  
		 &
		 0
		\end{bmatrix}\T.
	\end{align*}

	\medskip
		
\begin{figure}
	 \centering
	 \includegraphics[width=0.5\columnwidth]{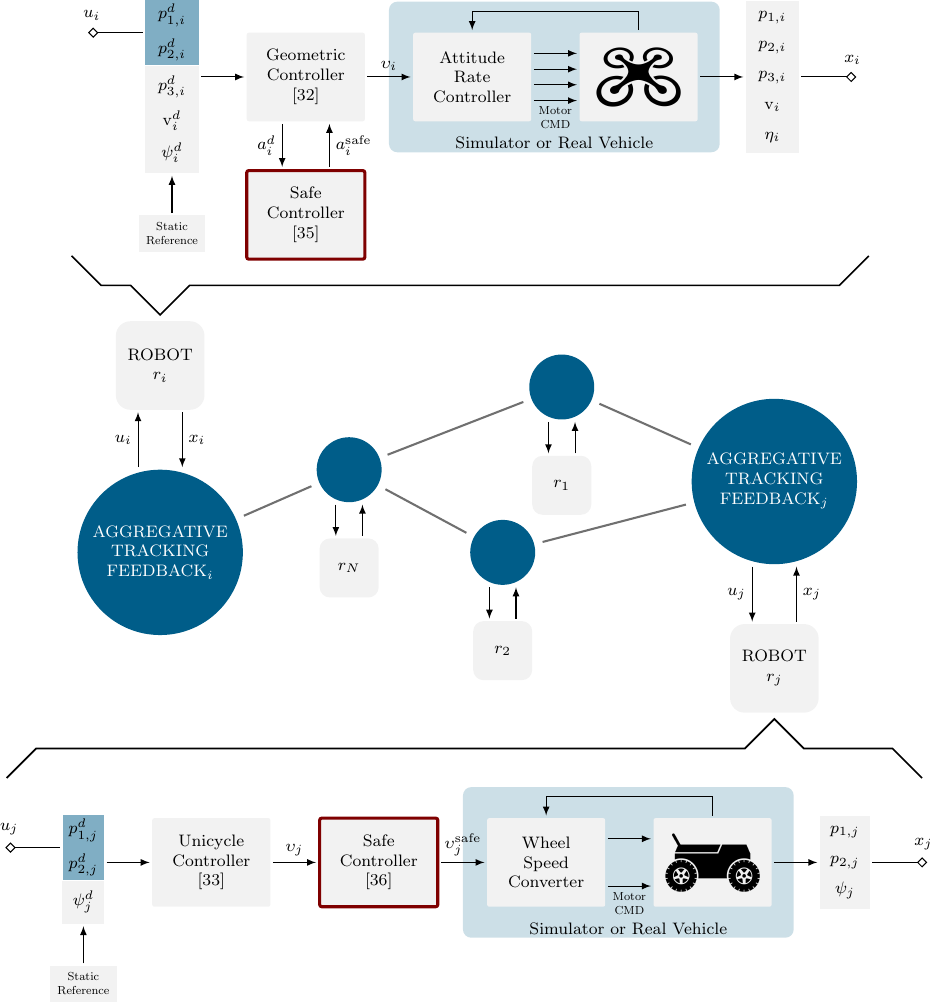}
	 \caption{
			Block diagram scheme illustrating the overall distributed architecture
			specialized for both the quadrotor and the ground mobile robot systems.
	 } 
	 \label{fig:plant}
	\end{figure}

	We present the overall system architecture in Fig.~\ref{fig:plant}. 
	Each node represents a robot within the heterogeneous team. 
	Specifically, in each node, we graphically emphasize the
        interconnection between the low-level (local) robot control
        architecture and the high-level (distributed) computing unit
        that, through communication with the other robots, implements
        \algo/.
      As for the low-level architecture, the upper part of the figure
      provides a detailed block diagram for a quadrotor, while the
      bottom part illustrates the same for a ground mobile robot.
      The ``Safe Controller'' blocks in Fig.~\ref{fig:plant} implement
      a mechanism based on CBFs~\cite{ames2016control} that we incorporate in
      our strategy to avoid collisions among the robots.
	More in detail, 
      we apply the CBF-based approach
      from~\cite{borrmann2015control} for collision avoidance among
      the quadrotors, and the one proposed
      in~\cite{wilson2020robotarium} for the team of ground robots.
	  Respectively, they generate the safe desired accelerations $a_{i}^{\text{safe}}\in\R^3$ for the geometric controller of the quadrotor, and, the safe control reference $\upsilon_j^\text{safe}\in\R^2$ for the wheel speed converter of the ground robot.
      The entire control architecture can be implemented by using the
      Robotic Operating System (ROS)~2, which provides tools for
      managing communication among robots and functionalities to
      realize (distributed) control stacks in a modular fashion.
      Further details on the implementation are provided in Section~\ref{sec:simulations}.

	\section{Virtual Experiments Via Robotic Simulator}
	\label{sec:simulations}
	
	In this section, we show the effectiveness of our distributed strategy considering a team of heterogeneous robots, i.e., aerial and ground robots, addressing the multi-robot target monitoring and encirclement task described in Section~\ref{sec:modeling}.

	Specifically, the aerial team is composed of Crazyflie nano-quadrotors, while the ground team is composed of TurtleBot $3$ Burger.
	These virtual experiments are performed using \textsc{CrazyChoir}~\cite{pichierri2023crazychoir}, a novel ROS~2 framework tailored for Crazyflies applications combined and based on \textsc{ChoiRbot}~\cite{testa2021choirbot}, a ROS~2 toolbox for Distributed Robotics.
	Here, each robot is governed by an independent ROS~2 control stack and supervised by a decentralized guidance layer that implements the distributed algorithm. 
	To this end, DISROPT~\cite{farina2020disropt}, a toolbox for distributed optimization, is exploited to implement the distributed algorithm on aerial and ground robots. 
	In fact, each robot communicates with a few neighbors according to the network communication graph.
	Moreover, \textsc{CrazyChoir} provides a virtual environment for swarm of nano-quadrotors based on Webots~\cite{michel2004cyberbotics}. 
	The same tool has been used to virtualize the TurtleBot $3$ Burger, employing the control stack provided by \textsc{ChoiRbot}.
	In the following, we show how the different contribution of the cost function in~\eqref{eq:cost_fcn} affects the team behavior.
	To better observe these dynamics, we consider only the aerial team for the first three scenarios, while we consider both aerial and ground robots in the last ones.
	The communication network of the robot is randomly generated according to an \er/ graph with edge probability $0.5$.
	We empirically set the tuning parameters of \algo/ according to $\alpha_1 = 0.5$, $\alpha_2 = 0.01$, and $\lambda = 0.05$.

	\subsection{Scenario 1: Encirclement Task via Aerial Team}

\begin{figure}
		\centering
		\includegraphics[width=0.5\columnwidth]{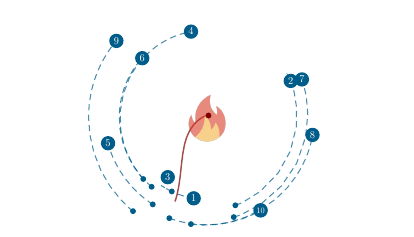}
		\caption{Top-down view of the Scenario 1 virtual experiment.
	 	The target is depicted by a fire. 
		The starting positions of the nano-quadrotors are depicted as small blue spheres, the final ones by bigger, numbered, spheres. 
		Here, the red line represent the trajectory of the barycenter of the team, while the blue dashed lines represent the trajectories of the nano-quadrotors.
		}
		\label{fig:encirclement_basic}
	\end{figure}

	First, we test our distributed strategy employing only the aerial team, composed by $N=10$ Crazyflies $2.1$. In this scenario, we aim to show the effectiveness of the cooperative target encirclement task, and, hence, we leverage only the first term of~\eqref{eq:cost_fcn}, namely
	\begin{align*}
		\f_i(x_i,\sigma(x)) = \f_{i}^{\text{coop}}(\sigma(x)),
	\end{align*}
	for all $i \in \until{N}$.
	In Fig.~\ref{fig:encirclement_basic}, we graphically show the role of $\f_i^{\text{coop}}(\agg(x))$ by providing the top-down view of a virtual experiment, conducted in a Webots environment, in which only this term has been considered.
	Here, we set the target in $b = \col(0,0)$.
	The team starts from a random position in the third quadrant of the plane, and, when the takeoff phase is completed, they cooperatively encircle the common wildfire area by running \algo/.
	It is important to note that the steady-state configuration is permutation-independent. 
	Specifically, since in this scenario the robots have to pursue only the cooperative, global objective (i.e., maximize the circular variance of the team around the target) without any individual local goal, they can start from any initial configuration and still surround the target without requiring a fixed, pre-assigned, angular set point.
	Here, the nano-quadrotors fly at the same height, encircling the common target avoiding mutual collisions.
	
	\begin{figure}
		\centering
		\includegraphics[width=0.5\columnwidth]{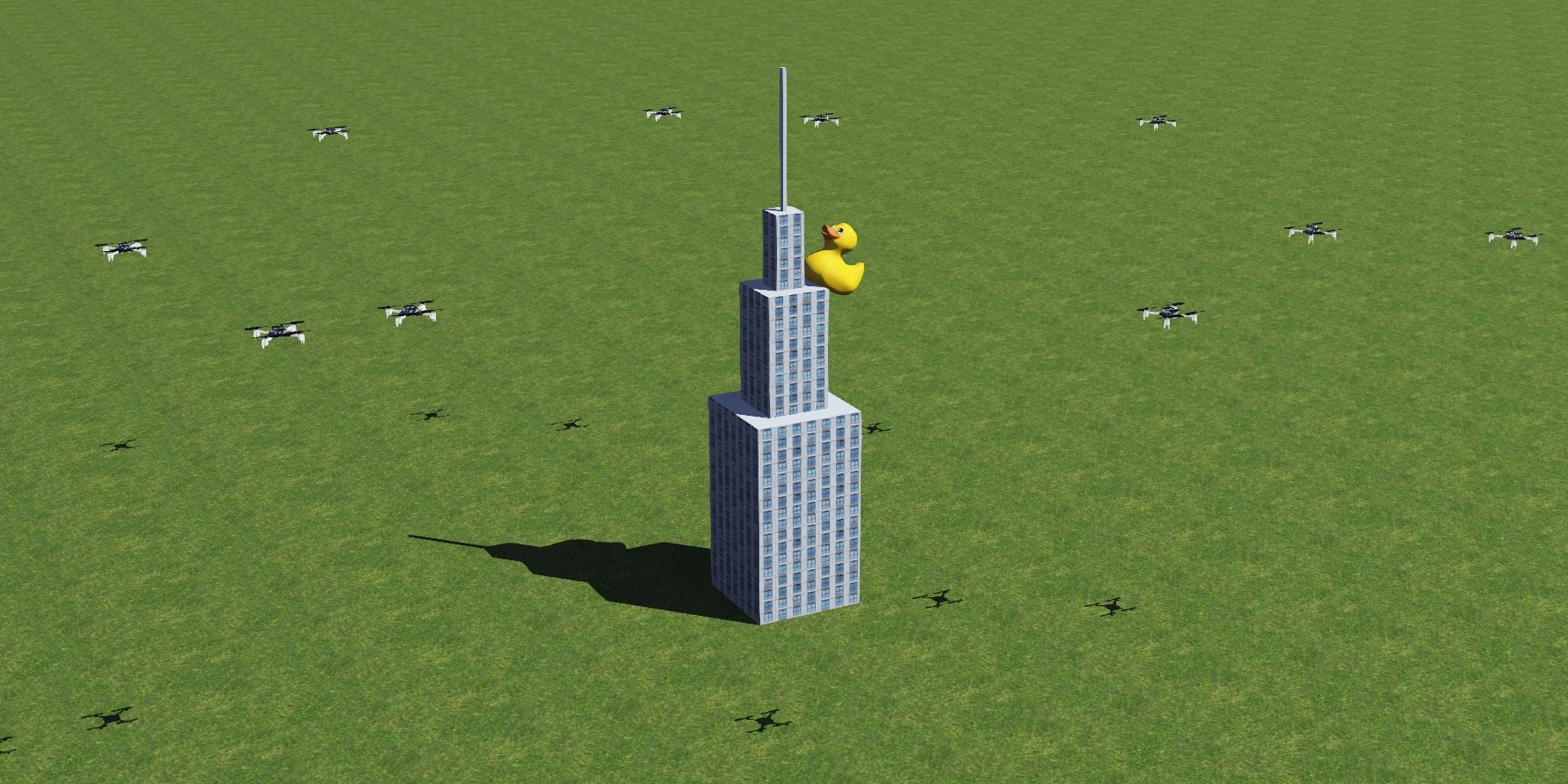}
		\caption{Snapshot from Webots of the final position of the Crazyflies in the virtual experiment from Scenario 1. The target is represented by a tower.}
		\label{fig:snapshot_webots}
	\end{figure}

	\subsection{Scenario 2: Multi-Spot Monitoring via Aerial Team}
	\label{subsec:encirclement_patrolling}

\begin{figure}
		\centering
		\includegraphics[width=0.5\columnwidth]{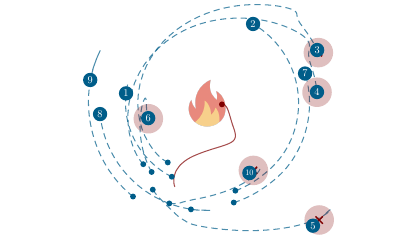}
		\caption{
		Top-down view of the Scenario 2 virtual experiment.
		As Fig.~\ref{fig:encirclement_basic}, the target is depicted by a fire, the starting locations by small spheres and the final by bigger, numbered spheres.
		The sensitive spot to monitor are the red circles.
		Crazyflies' trajectories are represented by blue dashed lines, while the barycenter trajectory is depicted by a red line.
		}
		\label{fig:encirclement_patrolling}
	\end{figure}

	In this second scenario, we also consider the term $\f_i^{\text{spot}}(x_i)$ thus embedding objective \textit{O2}, i.e., the need for multi-spot monitoring, namely
	\begin{align*}
		\f_i(x_i,\sigma(x)) = \gamma_1 \f_i^{\text{coop}}(\sigma(x)) + \gamma_2 \f_i^{\text{spot}}(x_i).
	\end{align*}
	In particular, in this scenario, we consider $\np=5$ points of interest to protect, which have been randomly assigned to $5$ robots.
	The planar coordinates $\point$ are randomly generated by extracting them with a uniform probability from the set $[-4,4]\times [-4,4]$, while the tuning parameters are set to $\gamma_1 = 25$ and $\gamma_2 = 2$, in order to prioritize the cooperative target encirclement.
	In Fig.~\ref{fig:encirclement_patrolling} we show the top-down view of the considered virtual experiment.
	In particular, Fig.~\ref{fig:encirclement_patrolling} shows that the steady-state configuration of the team is different from the one of the previous scenario since they now patrol sensitive points (\textit{local} task) while concurrently encircle the common target (\textit{global} task).

	\begin{figure}
		\centering
		\includegraphics[width=0.5\columnwidth]{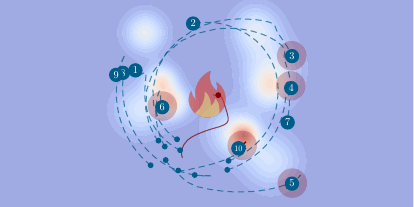}
		\caption{Top-down view of Scenario 3. 
		Here, the dangers' locations are depicted in the heatmap. }
		\label{fig:encirclement_wildfire}
	\end{figure}
	
	\subsection{Scenario 3: Safe Placement via Aerial Team}
	\label{subsec:encirclement_wildfires}

	In this section, we describe the complete scenario, including also the requirement represented by objective \textit{O3}, i.e., avoid the dangerous surroundings.
	Thus,
	we consider the overall cost in~\eqref{eq:cost_fcn}, i.e., involving also the cost function component $\gamma_3 \f_i^{\text{danger}}(x_i)$ with $\gamma_3 = 20$.
	The dangerous areas are modeled as a sum of $\nd = 10$ Gaussian functions $\mathcal{G}_l(x_i,\mu_l,\sigma_l)$, namely
	\begin{align*}
		\f_i^{\text{danger}}(x_i) = \sum_{\ell =1}^{\nd}\mathcal{G}_l(x_i,\mu_l,\sigma_l),
	\end{align*}
	with $\mu_l$ the mean of the $\ell$-th Gaussian function randomly generated with a uniform probability from the set $[-3.0, 3.0]$, and $\sigma_l$ the standard deviation of the $\ell$-th Gaussian function randomly generated with a uniform probability from the set $[0.4,0.9]$.
	In Fig.~\ref{fig:encirclement_wildfire}, we show the results of a final virtual experiment in which the whole team changed their positioning based on the heatmap generated by the sum of the Gaussian functions.
	In particular, the robots still cooperatively encircle the common target and position themselves near the points of interest, but now they concurrently avoid dangerous areas, moving themselves towards safe locations.
	In fact, in this scenario, both the trajectories and the final locations of some robots (e.g., the ones that do not monitor sensitive points) are different from the previous ones, as they avoid dangerous areas.

	\subsection*{Scenario 4 - Heterogeneous Team}
	\label{sec:hetero_sim}
	
	\begin{figure}
		\centering
		\includegraphics[width=0.5\columnwidth]{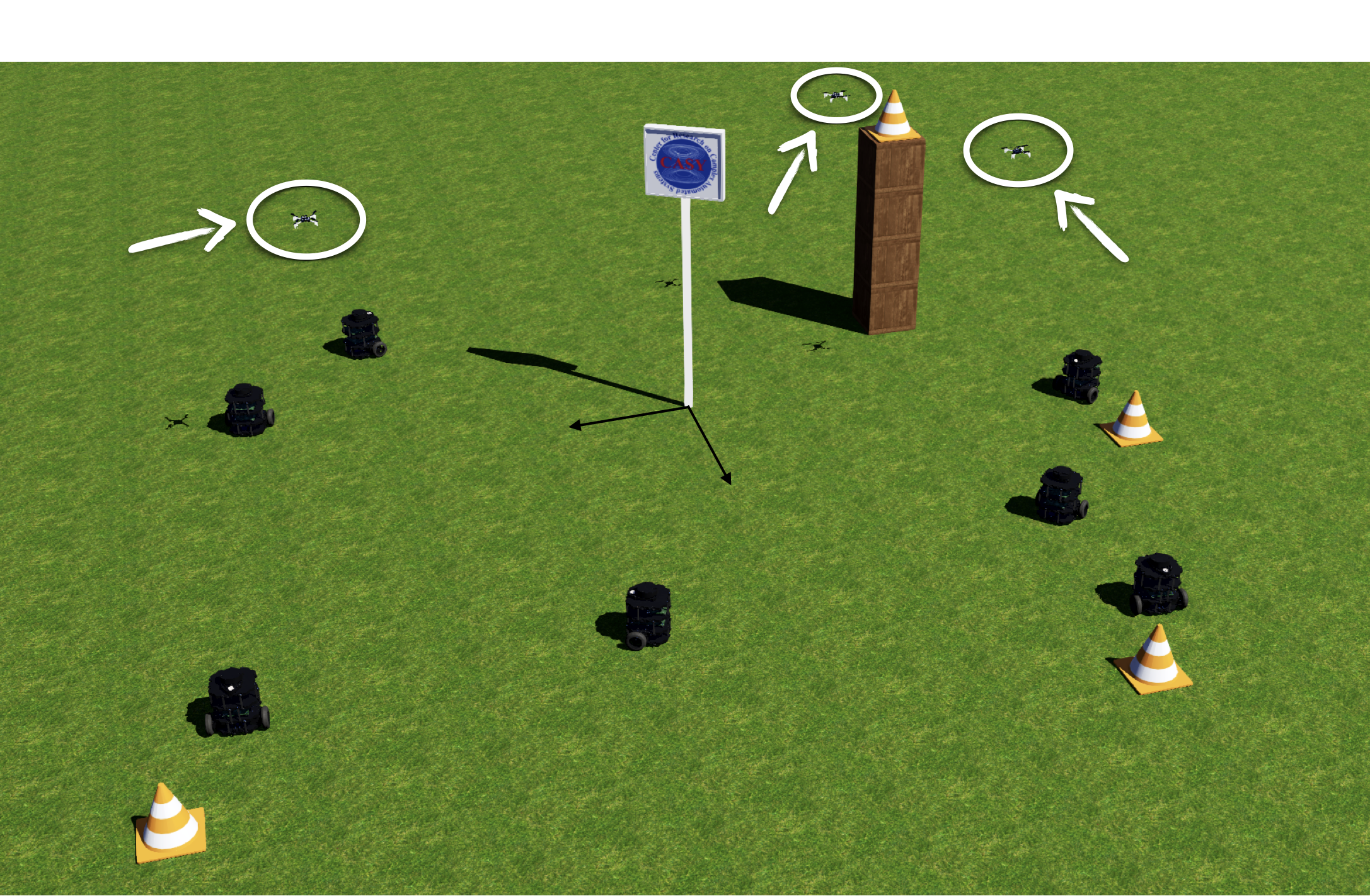}
		\includegraphics[width=0.5\columnwidth]{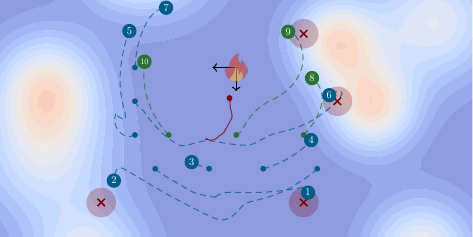}
		\caption{
		The above image captures a final instant from the Webots virtual experiment described in Scenario 4. 
		Here, a white signal serves as the target for encirclement, whereas the traffic cones highlight the sensitive spots to monitor. Below, we provide a top-down view of the trajectories of the robots.}
		\label{fig:encirclement_hetero}
	\end{figure}
	
	\begin{figure}
		\centering
		\includegraphics[width=0.5\columnwidth]{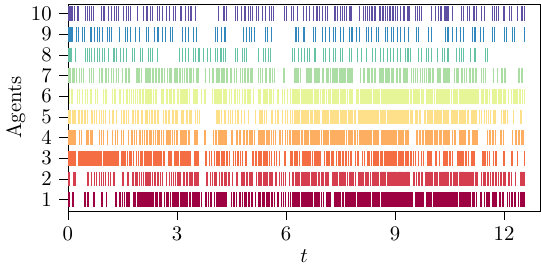}
		\caption{Occurrence of the triggering conditions during the heterogeneous virtual experiment of Scenario 4.}
		\label{fig:ET_encirclement_patrolling}
	\end{figure}

	We now consider a heterogeneous fleet of $N = 10$ robots, namely, $N_a = 3$ Crazyflie nano-quadrotors and $N_g = 7$ TurtleBot. 
	This team focuses on $S = 4$ points of interest to monitor.
	Specifically, one point is shared between two aerial robots, while the remaining three points are randomly assigned to the ground robots. 
	We show the results of this (heterogeneous) virtual experiment in Fig.~\ref{fig:encirclement_hetero}, where both aerial robots (in green) and ground ones (in blue) cooperate to encircle the common target placed in $b=\col(0,0)$. 
	Here, aerial robot $9$ moves toward its point of interest, as well as the ground robots $1,2,6$. 
	Indeed, the other members of the team balance the encirclement task, embracing the target. 
	Finally, every robot successfully avoids dangerous zones, achieving the last task of the designed mission. 
	From this virtual experiment, it is way more visible the action of collision avoidance among the robots. 
	In fact, robot $1$ does not collide with robot $2$ but reaches its steady-state location successfully. 
	Moreover, collision avoidance contributes also to robot $6$, which, in the middle of its trajectory, avoids robot $3$. 
	We finally remark that the aerial couple (see the first quadrant of the picture) never reaches a collision.
	
	Moreover, in Fig.~\ref{fig:ET_encirclement_patrolling}, we graphically show the effects of the employed event-triggered communication strategy.
	Indeed, the lines in Fig.~\ref{fig:ET_encirclement_patrolling} represent the occurred communications for each of the nano-quadrotors running \algo/ with $\lambda = 0.05$.
	The plot demonstrates how event-triggered communication effectively reduces inter-agents communication. In fact, the nano-quadrotors exchange data less frequently, only when needed, i.e., when the condition in \eqref{eq:triggering_law} is satisfied.

	A video is available as supplementary material to the
	paper\footnote{\label{video}The video is available at
	\url{https://youtu.be/iIUChcNUdr4}}.

	\subsection{Monte Carlo Virtual Experiments}
	\label{subsec:monte_carlo}
	We performed two Monte Carlo virtual experiments on random instances of problem~\eqref{eq:aggregative_problem} on the realistic simulator Webots with a heterogeneous team of robots.
	We test the performance of the proposed algorithm with different values of the number of robots $N$ and the triggering communication parameter $\lambda$.
	
	\subsubsection*{First Virtual Experiment}

\begin{figure}
		\centering
		\includegraphics[width=0.5\columnwidth]{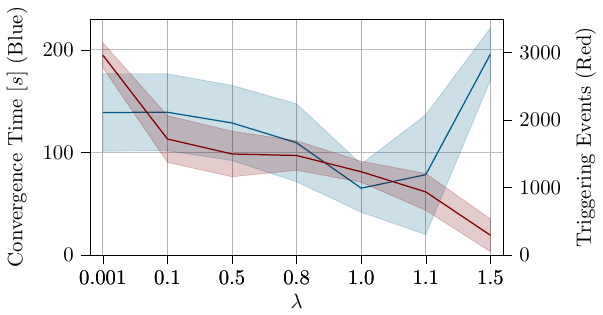}
		\caption{Algorithm performance increasing the triggering communication parameter $\lambda$. Blue: convergence time. Red: number of triggering events. The shaded areas represent the standard deviation band. 
		}
		\label{fig:montecarlo_lambda}
	\end{figure}

	To begin with, we test the effectiveness of the algorithm by analyzing the time needed to reach a stationary point of the problem while varying the triggering communication parameter $\lambda$. 
	We perform $50$ Monte Carlo trials for each value of $\lambda$ and we fix the number of nano-quadrotors to $N=5$. 
	For each trial, the initial positions of the robots are randomly generated on the plane, along with the communication graph $\cG$. 
	We consider the cooperative target encirclement task, and, hence, we leverage only the first term of~\eqref{eq:cost_fcn}.
	Moreover, to focus the analysis on the effectiveness of the algorithm, we make the nano-quadrotors fly at different altitudes, and, hence, we do not make use of the safety control barrier functions to avoid collision.
	In Fig.~\ref{fig:montecarlo_lambda}, we show how the triggering communication parameter $\lambda$ affects the algorithm performance.
	Specifically, we demonstrate that the algorithm can significantly reduce the number of communication events (in red) without affecting its convergence time (in blue), up to an optimal value of $\lambda$. 
	In each trial, the convergence time is defined as the time required to achieve $\norm{\nabla\fsh(u)} \leq 5\cdot10^{-2}$, while the triggering events represent the number of communications over the number of agents.
	As shown in Fig.~\ref{fig:montecarlo_lambda}, in this set of virtual experiments, the optimal range for $\lambda$ is between $1.0$ and $1.1$.

	\subsubsection*{Second Virtual Experiment}

	For the second Monte Carlo virtual experiment, we analyze the scalability of the algorithm while varying the number of robots. 
	We perform $50$ Monte Carlo trials for each value of $N$ and we fix the triggering communication parameter to $\lambda=1.0$.
	For each trial, the initial positions of the robots are randomly generated on the plane, along with the communication graph $\cG$.
	In this set of virtual experiments, we test the cooperative target encirclement task over a fleet of ground robots, and we leverage only the first term of~\eqref{eq:cost_fcn}. 
	Here, we also include the safety CBFs to avoid direct collisions among the robots.

\begin{figure}
		\centering
		\includegraphics[width=0.5\columnwidth]{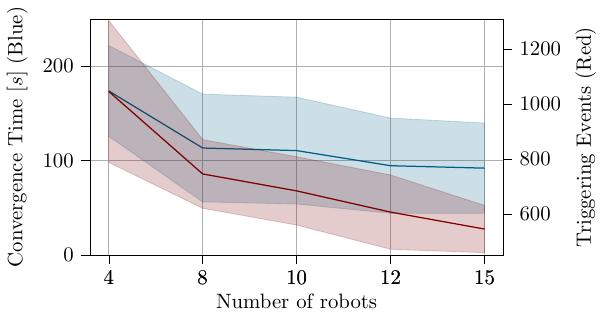}
		\caption{Algorithm scalability increasing the number of robots. Blue: convergence time. Red: communication events occurrences normalized on the number of robots. The shaded areas represent the standard deviation band.
		}
		\label{fig:montecarlo_N}
	\end{figure}

	In Fig.~\ref{fig:montecarlo_N}, we show the results of the second Monte Carlo virtual experiment. 
	As before, in each trial, the convergence time is defined as the time required to achieve $\norm{\nabla\fsh(u)} \leq 1\cdot10^{-3}$, while the triggering events represent the number of communications normalized on the number of agents.
	Here, the algorithm scalability is confirmed as the convergence time (in blue) decreases, while the communication events occurrences (in red) drops significantly.

	\section{Experiments With Nano-quadrotors and Ground Mobile Robots}
	\label{sec:experiments}
	
	To confirm the robustness of our solution, we show experimental results on real fleets of robots that, in a fully-distributed fashion, perform the encirclement of a common target while concurrently monitoring sensitive points of interest and avoiding dangerous areas.

	\subsection{Experimental Setup}

	The heterogeneous fleet is composed of $N = 7$ robots, $N_a = 3$ Crazyflie nano-quadrotors, and $N_g = 4$ TurtleBot.
	To physically control the Crazyflies, we sent via radio the angular-rate set points computed via the geometric nonlinear controller provided by \textsc{CrazyChoir}.
	In fact, Crazyflies are equipped with an nRF51822 radio controller, used to communicate with the local workstation by means of the Crazyradio PA, a 2.4 GHz USB radio dongle based on the nRF24LU1+ sensor. 
	On the other hand, TurtleBot $3$ burger requires a control input in both linear and angular velocity, generated by the control libraries provided by \textsc{ChoiRbot}. 
	Here, the ground fleet communicates with the workstation leveraging the well-known ROS~2 topic-subscriber protocol, which relies on a standard 2.4GHz WiFi network.
	The whole real and virtual experiments have been dockerized and run on an Ubuntu 20.04 workstation equipped with Intel i9 processor and an Nvidia RTX 4000 GPU. 
	The ROS~2 version used in our experiment is Foxy Fitzroy.
	The pose of the robots (i.e., position and orientation) is provided by the Vicon MoCap System. 
	In fact, a second workstation (Intel Xeon, Windows 7) handles the Vicon Tracker software.

	\subsection{Experiment with a Heterogeneous Fleet}

	We now present a real experiment of the proposed solution with a heterogeneous fleet of robots. 
	Here, we consider a heterogeneous team of $N = 7$ robots, $N_a = 3$ Crazyflie nano-quadrotors, and $N_g = 4$ TurtleBot.
	Here, we set $S = 4$ sensitive points to monitor.
	Specifically, two points have been assigned to the aerial robots, while the remaining two points are assigned to the ground robots.

\begin{figure}
		\centering
		\includegraphics[width=0.5\columnwidth]{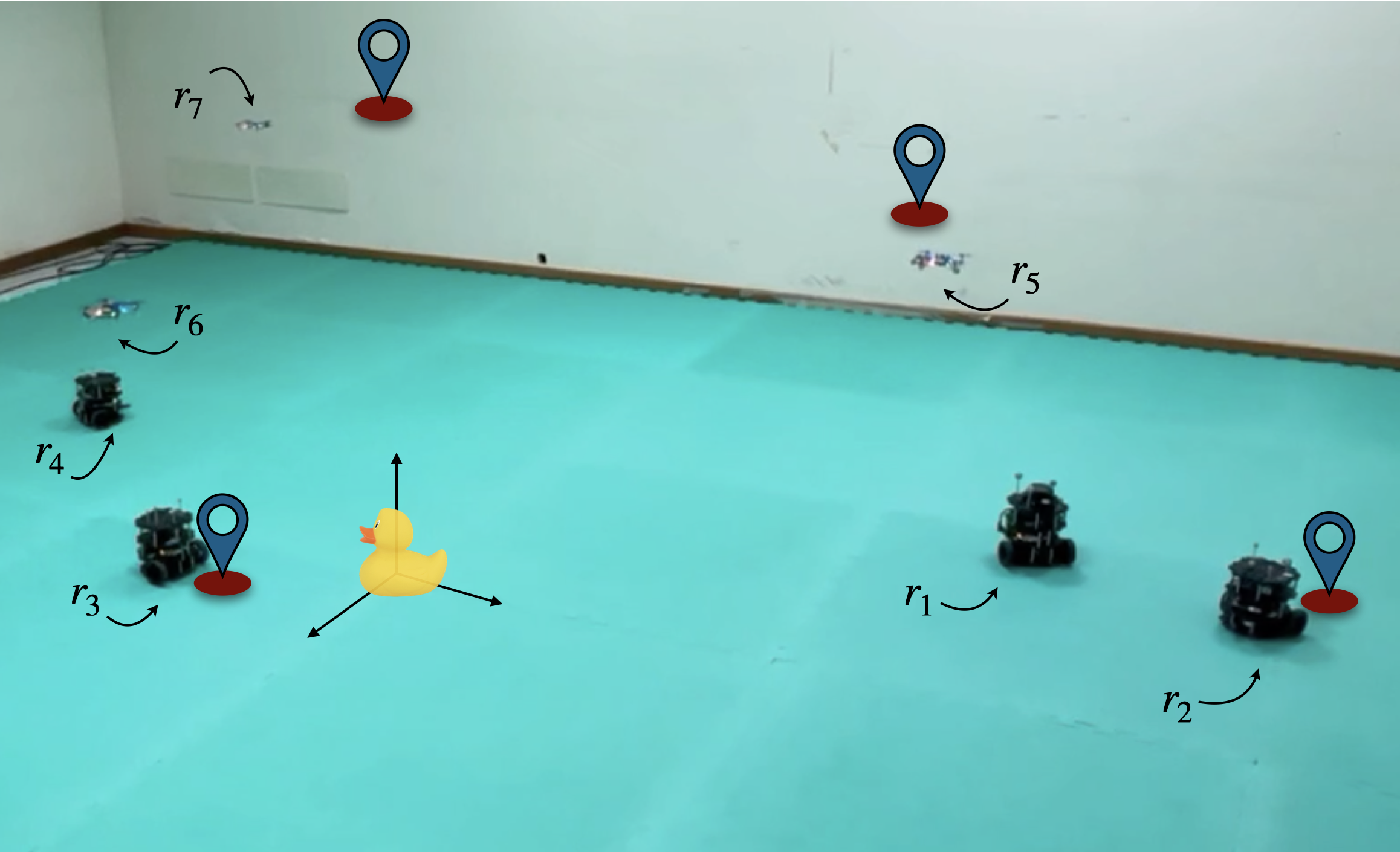}
		\includegraphics[width=0.5\columnwidth]{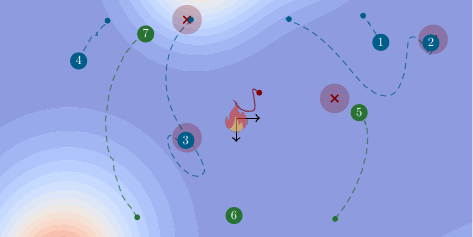}
		\caption{
		The image above provides the final instant of the experiment conducted in our laboratory. 
		A small yellow duck serves as the target for encirclement, whereas the pin-points highlight the points of interest to monitor. 
		The image below provides a top-down view of the robots' trajectories.
		}
		\label{fig:experiment}
	\end{figure}

	We present the results of this experiment in Fig.~\ref{fig:experiment}, where both aerial robots (in green) and ground ones (in blue) cooperate to encircle the common target placed in $b=\col(0,0)$. 
	Here, it is interesting to note how robot $6$ stays steady in the middle of the plane, clearly balancing the positioning of the team. On the other hand, the nano-quadrotor $5$ and $7$ move toward their points of interest, as well as the ground robot $2$.
	In particular, it is worth mentioning that this Turtlebot (robot $2$) does not collide with robot $1$, drawing a trajectory that avoids it.
	Indeed, the other members of the team balance the encirclement task, embracing the target. 
	The last remark regards robot $7$, which remains at a distance from its personal point of interest because placed in dangerous areas.
	A video\footnotemark[1] is available as supplementary material to the
	paper.

	\section{Conclusion}
	\label{sec:Conclusions}
	In this paper, we proposed a distributed feedback optimization
        strategy to address multi-robot target monitoring
        and encirclement by leveraging the aggregative feedback
        optimization framework.
	We designed a distributed feedback optimization policy
        incorporating a triggered communication mechanism based on
        locally verifiable conditions. We demonstrated that the
        resulting asynchronous, distributed policy successfully drives
        the robots to a configuration satisfying first-order necessary
        conditions of optimality.
	We implemented the proposed strategy into a ROS~2 distributed
        control architecture and validated it through realistic
        virtual experiments using the Webots robotic simulator. The
        virtual experiments include a Monte Carlo campaign
        highlighting the communication load and the scalability of the
        strategy. Furthermore, we conducted experiments on a real
        testbed featuring a heterogeneous fleet of ground and aerial
        robots, providing practical evidence of the solution
        applicability.

	\section*{Acknowledgments}
	The authors would like to deeply thank Alice Rosetti for her support
        in the development of the experiment.

	\appendix 
	
	\subsection{Proof of Theorem~\ref{th:convergence}}
	\label{sec:proof}

	The result~\cite[Th.~3.1]{carnevale2024nonconvex} assesses the attractiveness and stability properties for the version of system~\eqref{eq:local_closed_loop} using a continuous-time communication instead of the event-triggered one ruled by~\eqref{eq:triggering_law}.
	Hence, we will use~\cite[Th.~3.1]{carnevale2024nonconvex} as a building block to obtain our proof.
	Indeed, we interpret the networked dynamics arising from~\eqref{eq:local_closed_loop} as the perturbed version of the nominal system studied in~\cite[Th.~3.1]{carnevale2024nonconvex}.
	More in detail, the perturbation is due to the triggering condition~\eqref{eq:triggering_law} and, thanks to the peculiar design of this condition, can be bounded in norm with a term vanishing in the case of configurations corresponding to stationary points of problem~\eqref{eq:aggregative_problem}. 

	After this preliminary introduction, let us add and subtract $\dfrac{1}{\alpha_2}a_{ij}(\w_i + \phii(\x_i) - \w_j - \phi_j(\x_j))$ and $\dfrac{1}{\alpha_2}a_{ij}(\z_i + \nabla_2 \f_i(\x_i,\w_i + \phii(\x_i)) - \z_j - \nabla_2 \f_j(\x_j,\w_j + \phi_j(\x_j)))$ into~\eqref{eq:local_closed_loop_w_i} and~\eqref{eq:local_closed_loop_z_i}, respectively.
	By resorting to the definition of each $\ei$ (cf.~\eqref{eq:e_i}), we rewrite~\eqref{eq:local_closed_loop_w_i} and~\eqref{eq:local_closed_loop_z_i} as
	\begin{subequations}\label{eq:local_closed_loop_eu}
		\begin{align}
			\dot{\w}_i &= -\dfrac{1}{\alpha_2}\sum_{j\in\cN_i}a_{ij}\left(\w_i + \phii(\x_i) - \w_j - \phi_j(\x_j)\right)
			- \dfrac{1}{\alpha_2} \sum_{j\in\cN_i} a_{ij}  \left(\ewi  -  \ewj\right)
			\label{eq:local_closed_loop_w_i_e_i}
			\\
			\dot{\z}_i &=
			-\dfrac{1}{\alpha_2}\sum_{j\in\cN_i}  a_{ij}(\z_i  +  \nabla_2 \f_i(\x_i,\w_i + \phii(\x_i)))
			+\dfrac{1}{\alpha_2}\sum_{j\in\cN_i}  a_{ij} (\z_j + \nabla_2 \f_j(\x_j,\w_j + \phi_j(\x_j)))
			-\dfrac{1}{\alpha_2}\sum_{j\in\cN_i}  a_{ij}(\ezi - \ezj)
			,\label{eq:local_closed_loop_z_i_e_i}
		\end{align}
	\end{subequations}
	for all $t \in [\btk,\btkp)$, which leads to the networked dynamics 
	\begin{subequations}\label{eq:closed_loop_with_bar}
		\begin{align}
			\dot{\x} &= \dyn(\x,\uu)
			\\
			\du  &= - \alpha_1\dir(\uu,\x,\w,\z)
			\\
			\dot{\w} &= -\dfrac{1}{\alpha_2}L\w-\dfrac{1}{\alpha_2}L\phi(\x) - \dfrac{1}{\alpha_2}L\ew\label{eq:dot_w}
			\\
			\dot{\z} &= -\dfrac{1}{\alpha_2}L\z-\dfrac{1}{\alpha_2}L\Gt(\x,\w) - \dfrac{1}{\alpha_2}L\ez\label{eq:dot_z}
			\\
			\dot{\xi} &= -\nu\xi,
		\end{align}
	\end{subequations}
	where $L \coloneqq \cL \otimes I_d \in \R^{Nd \times Nd}$, $\ew\coloneqq \col(\ew_1,\dots,\ew_N)$, $\ez \coloneqq \col(\ez_1,\dots,\ez_N)$, $\dir(\x,\uu,\w,\z) \coloneqq \col(\dir_1(\x_1,\uu_1,\w_1,\z_1),\dots,\dir_N(\x_N,\uu_N,\w_N,\z_N))$, $\phi(\x) \coloneqq \col(\phi_1(\x_1),\dots,\phi_N(x_N))$, and $\Gt(\x,\w) \coloneqq \col(\nabla_2 \f_1(\x_1,\w_1 + \phi_1(\x_1)),\dots,\nabla_2 \f_N(\x_N,\w_N + \phi_N(\x_N)))$.
	Then, inspired by~\cite{carnevale2024nonconvex}, we introduce $R \in \R^{Nd \times (N-1)d}$ such that $R\T \1 = 0$ and $R\T R = I$ and the new variables $\bz, \bw \in \R^{d}$ and $\pw, \pz \in \R^{N(d-1)}$ defined as 
	\begin{align*}
		\begin{bmatrix}
			\w
			\\
			\z
		\end{bmatrix} \longmapsto 
		\begin{bmatrix}
			\bw
			\\
			\pw
			\\
			\bz 
			\\
			\pz
		\end{bmatrix} \coloneqq
		\begin{bmatrix}
			\1\T \w
			\\
			R\T \w
			\\
			\1\T \z 
			\\
			R\T \z
		\end{bmatrix}.
	\end{align*}
	In light of Assumption~\ref{ass:network}, it holds $\1\T L = 0$ and, thus, we get $\1\T \dot{w} = \1\T \dot{z} = 0$.
	Hence, since $\1\T\w(0) = \1\T\z(0) = 0$ by assumption, it holds $\1\T\w(t) = \1\T\z(t) = 0$ for each $t \ge 0$ and, thus, we neglect them from the analysis rewriting~\eqref{eq:closed_loop_with_bar} as%
	\begin{subequations}\label{eq:closed_loop}
		\begin{align}
			\dot{\x} &= \dyn(\x,\uu)\label{eq:dot_x}
			\\
			\du  &= - \alpha_1\dir(\uu,\x,R\pw,R\pz)\label{eq:dot_u}
			\\
			\dot{\w}_\perp &= -\dfrac{1}{\alpha_2}R\T LR\pw  -  \dfrac{1}{\alpha_2}R\T L\phi(\x)  -  \dfrac{1}{\alpha_2}R\T L\ew\label{eq:dot_pw}
			\\
			\dot{\z}_\perp &= -\dfrac{1}{\alpha_2}R\T LR \pz-\dfrac{1}{\alpha_2}R\T L\Gt(\x,R\pw) - \dfrac{1}{\alpha_2}R\T L\ez\label{eq:dot_pz}
			\\
			\dot{\xi} &= -\nu\xi.
		\end{align}
	\end{subequations}
	Since $RR\T = I - \frac{\1\1\T}{N}$, $R\T\1 = 0$, $\frac{\1\T}{N}\phi(\x) = \sigma(\x)$, and $\1\T\Gt(\x,\1\sigma(\x)) = \sum_{j=1}^N \nabla \f_j(\x_j,\sigma(\x))$, we note that 
	\begin{align}
		&\dir(\x,\uu,-RR\T\phi(\x),-RR\T\Gt(\x,\1\sigma(\x))) 
		&=\dir(\x,\uu,\1\sigma(\x),\tfrac{\1\1\T}{N}\Gt(\x,\1\sigma(\x)))
		&= \nabla h(\uu)\nabla \fs(\x),\label{eq:dir_steps}
	\end{align}
	namely, if $\pw = - R\T\phi(\x)$ and $\pz = - R\T\Gt(\x,\1\sigma(\x))$, the direction $\nabla h(\uu)\nabla \fs(\x)$ is available in the dynamics of $\uu$.
	Then, by following the same steps in the proof of~\cite[Th.~3.1]{carnevale2024nonconvex}, we introduce the error coordinate $\rr \in \R^{2(N-1)d}$ defined as 
	\begin{align}\label{eq:rr}
		\begin{bmatrix}
			\x 
			\\
			\pw 
			\\
			\pz
		\end{bmatrix}  \longmapsto 
		\begin{bmatrix}\x 
			\\
			\rr
		\end{bmatrix}  \coloneqq  \begin{bmatrix}\x 
			\\
			\tpw 
			\\
			\tpz
		\end{bmatrix}  \coloneqq  \begin{bmatrix}
			\x 
			\\
			\pw + R\T\phi(\x)
			\\
			\pz + R\T\Gt(\x,\1\sigma(\x))
		\end{bmatrix}.
	\end{align}
	Let us introduce $\chi \coloneqq \col(\x,\uu,\rr) \in \R^{\nchi}$ with $\nchi = n + m + 2N(d-1)$ and $\chi \in \R^{\nchi}$ and rewrite~\eqref{eq:closed_loop} as 
	\begin{subequations}\label{eq:perturbed}
		\begin{align}
			\dot{\chi} &= \nom(\chi) + \e
			\\
			\dot{\xi} &= -\nu\xi
		\end{align}
	\end{subequations}	
	where $\nom: \R^{\nchi} \to \R^{\nchi}$ compactly collects the dynamics in~\eqref{eq:dot_x}--\eqref{eq:dot_pz} written in the new coordinates (see~\cite{carnevale2024nonconvex} for the explicit definition), while $\e \in \R^{\nchi}$ reads as 
	\begin{align*}
		\e \coloneqq - \dfrac{1}{\alpha_2}\begin{bmatrix}
			0
			\\
			R\T L\ew 
			\\
			R\T L\ez 
		\end{bmatrix}.
	\end{align*}
	By using the triangle inequality, the Cauchy-Schwarz inequality, and the definition of $\e$, we get
	\begin{align}
		\norm{\e}
		&\leq \dfrac{1}{\alpha_2}\norm{R\T L} \sum_{i=1}^N \left(\norm{\ew_i}  + \norm{\ez_i}\right)
		\notag\\
		&
		\stackrel{(a)}{\leq} 
		\tr\dfrac{1}{\alpha_2}\norm{R\T L} \sum_{i=1}^N \norm{\diri(\ui,\x_i,\w_i,\z_i)}
		+\tr\dfrac{1}{\alpha_2}\norm{R\T L}\sum_{i=1}^N\norm{\xi_i}
		\notag\\
		&\stackrel{(b)}{=} 
		\tr\dfrac{1}{\alpha_2}\sqrt{N}\norm{R\T L}\norm{\dir(\uu,\x,\w,\z)}
		+\tr\dfrac{1}{\alpha_2}\sqrt{N}\norm{R\T L}\norm{\xi}
		\notag\\
		&\stackrel{(c)}{=} 
		\tr\dfrac{1}{\alpha_2}\sqrt{N}\norm{R\T L}\norm{\dir(\uu,\x, R\pw, R\pz)}
		+\tr\dfrac{1}{\alpha_2}\sqrt{N}\norm{R\T L}\norm{\xi},\label{eq:bound_err_intermediate}
	\end{align}
	where in $(a)$ we use the condition~\eqref{eq:triggering_law}, in $(b)$ we apply a basic algebraic property and the definitions of $\dir$ and $\xi$, while in $(c)$ we use $\w = R\pw$ and $\z = R\pz$.
	To further bound~\eqref{eq:bound_err_intermediate}, let $\tdir:  \R^{n}  \times  \R^{m}  \times  \R^{(N-1)d}  \times  \R^{(N-1)d}  \to  \R^{m}$ be defined as 
	\begin{align*}
		&
		\tdir(\x,\uu,\tpw,\tpz) 
		&=
		\dir\left(\uu,\x,\1\sigma(\x) + R\tpw,\tfrac{\1\1\T}{N}\Gt(\x,\1\sigma(\x)) + R\tpz\right)
		-\dir\left(\uu,\x,\1\sigma(\x),\tfrac{\1\1\T}{N}\Gt(\x,\1\sigma(\x))\right).
	\end{align*}
	Then, by relying on the steps in~\eqref{eq:dir_steps}, the new coordinates in~\eqref{eq:rr}, and the definitions of $g$, $\fs$, and $\fsh$, it holds
	\begin{align}
		\dir(\uu,\x,R\pw,R\pz)
		&
		= 
		\nabla h(\uu)\nabla\fs(\x) + \tdir(\x,\uu,\tpw,\tpz)
		\notag\\
		&
		\stackrel{(a)}{=}
		\nabla\fsh(\uu) + \nabla h(\uu)(\nabla\fs(h(\uu)) - \nabla\fs(\x)) 
		+ \tdir(\x,\uu,\tpw,\tpz),\label{eq:dir_equivalence}
	\end{align}
	where in $(a)$ we add $\pm\nabla h(\uu)\nabla \fs(h(\uu)) = \nabla \fsh(\uu)$.
	By using the Lipschitz property of each $h_i$, $\phii$, and $\nabla \f_i$ (cf. Assumption~\ref{ass:steady_state} and~\ref{ass:lipschitz}), we use~\eqref{eq:dir_equivalence} to get the bound 
	\begin{align*}
		\norm{\dir(\uu,\x,R\pw,R\pz)} 
		&\leq \norm{\nabla\fsh(\uu)} + \lipp_h\lipp_0^\agg\norm{\x - h(\uu)}
		+ \lipp_h(\lipp_1 + \lipp_2\lipp_3)\norm{\tpw} + \lipp_h\lipp_2\lipp_3\norm{\tpz},%
	\end{align*}
	which allows us to further bound~\eqref{eq:bound_err_intermediate} as
	\begin{align}
		\norm{\e}
		&\leq \tr \dfrac{1}{\alpha_2}c\norm{\begin{bmatrix}
			\nabla \fsh(\uu)
			\\
			\x - h(\uu)
			\\
			\rr
		\end{bmatrix}} + \dfrac{1}{\alpha_2}c_\xi\norm{\xi},\label{eq:bound_err}
	\end{align}
	where $c \coloneqq \sqrt{N}\norm{R\T L}\max\{1,\lipp_h\lipp_0^\agg,\lipp_h(\lipp_1 + \lipp_2\lipp_3),\lipp_h\lipp_2\lipp_3\}$ and $c_\xi \coloneqq \sqrt{N}\norm{R\T L}$.
	We recall that, under Assumptions~\ref{ass:network},~\ref{ass:steady_state}, and~\ref{ass:lipschitz}, the proof of~\cite[Th.~3.1]{carnevale2024nonconvex} guarantees the existence of a radially unbounded $V: \R^{\nchi} \to \R$ and $\bar{\alpha}_1, \bar{\alpha}_2, c_1, c_2 > 0$ such that, for any $\alpha_1 \in (0,\bar{\alpha}_1)$ and $\alpha_2 \in (0,\bar{\alpha}_2)$, it holds
	\begin{subequations}\label{eq:dot_V_conditions}
		\begin{align}
			\dfrac{\partial V(\chi)}{\partial \chi}\nom(\chi)  &\leq  - c_1
				\norm{\begin{bmatrix}
				\nabla \fsh(\uu)
				\\
				\x  -  h(\uu)
				\\
				\rr
			\end{bmatrix}}^2\label{eq:dot_V_conditions_minus}
			\\
			\norm{\dfrac{\partial V(\chi)}{\partial \chi}} &\leq c_2
				\norm{\begin{bmatrix}
				\nabla\fsh(\uu)
				\\
				\x  -  h(\uu)
				\\
				\rr
			\end{bmatrix}},\label{eq:dot_V_conditions_bound}
		\end{align}
	\end{subequations}
	for all $\chi, \chi^\prime \in \R^{\nchi}$.
	With these results at hand, we choose $\alpha_1 \in (0,\bar{\alpha}_1)$ and $\alpha_2 \in (0,\bar{\alpha}_2)$, we define $V_\xi(\chi,\xi) \coloneqq V(\chi) + \frac{1}{2}\norm{\xi}^2$, and, along the trajectories of~\eqref{eq:perturbed}, we get
	\begin{align}
		\dot{V}_\xi(\chi,\xi)  &=  \dfrac{\partial V(\chi)}{\partial \chi}(\nom(\chi) + \e) - \nu\norm{\xi}^2
		\notag\\
		&\stackrel{(a)}{\leq} 
		- c_1\norm{\begin{bmatrix}
			\nabla\fsh(\uu)
			\\
			\x - h(\uu)
			\\
			\rr
		\end{bmatrix}}^2
		+ \norm{ \dfrac{\partial V(\chi)}{\partial \chi}}\norm{\e} - \nu\norm{\xi}^2
		\notag\\
		&\stackrel{(b)}{\leq} 
		- c_1\norm{\begin{bmatrix}
			\nabla\fsh(\uu)
			\\
			\x - h(\uu)
			\\
			\rr
		\end{bmatrix}}^2
	+ \tr c_3\norm{\begin{bmatrix}
			\nabla\fsh(\uu)
			\\
			\x - h(\uu)
			\\
			\rr
		\end{bmatrix}}^2  
		+c_4\norm{\begin{bmatrix}
			\nabla\fsh(\uu)
			\\
			\x - h(\uu)
			\\
			\rr
		\end{bmatrix}}\norm{\xi} - \nu\norm{\xi}^2,\label{eq:dot_V}
	\end{align}
	where in $(a)$ we use~\eqref{eq:dot_V_conditions_minus} and the Cauchy-Schwarz inequality, while in $(b)$ we combine the bounds~\eqref{eq:bound_err} and~\eqref{eq:dot_V_conditions_bound} and introduce $c_3 \coloneqq cc_2\frac{1}{\alpha_2}$ and $c_4\coloneqq c_\xi c_2\frac{1}{\alpha_2}$.
	Now, let us arbitrarily choose $\tilde{c}_1 \in (0,c_1)$ and define $\bar{\tr} \coloneqq \frac{\tilde{c}_1-c_1}{c_3}$.
	Then, for any $\tr \in (0,\bar{\tr})$, we bound the right-hand side of~\eqref{eq:dot_V} as 
	\begin{align}
		\dot{V}_\xi(\chi,\xi) &\leq - \tilde{c}_1\norm{\begin{bmatrix}
			\nabla\fsh(\uu)
			\\
			\x - h(\uu)
			\\
			\rr
		\end{bmatrix}}^2 +c_4\norm{\begin{bmatrix}
			\nabla\fsh(\uu)
			\\
			\x - h(\uu)
			\\
			\rr
		\end{bmatrix}}\norm{\xi} 
		- \nu\norm{\xi}^2.\label{eq:dot_V_final}
	\end{align}
	Let us introduce 
	\begin{align*}
		\zeta(\x,\uu,\rr) \coloneqq 
		\begin{bmatrix}
			\nabla\fsh(\uu)
			\\
			\x - h(\uu)
			\\
			\rr
		\end{bmatrix}.
	\end{align*}
	Then, we rewrite~\eqref{eq:dot_V_final} in a matrix form as 
	\begin{align}
		\dot{V}_\xi(\chi,\xi)
			&\leq  -  \begin{bmatrix}
			\norm{\zeta(\x,\uu,\rr)}
			\\
			\norm{\xi}
		\end{bmatrix}\T  \begin{bmatrix}
			\tilde{c}_1& -c_4/2
			\\
			-c_4/2& \nu
		\end{bmatrix} \begin{bmatrix}
			\norm{\zeta(\x,\uu,\rr)}
			\\
			\norm{\xi}
		\end{bmatrix}.\label{eq:dot_V_matrix}
	\end{align}
	Let $\bar{\nu} \coloneqq \frac{c_4^2}{4\tilde{c}_1}$.
	Then, for any $\nu > \bar{\nu}$, the matrix in~\eqref{eq:dot_V_matrix} is positive definite and, thus, by denoting with $m > 0$ its smallest eigenvalue, we can further bound~\eqref{eq:dot_V_matrix} as 
	\begin{align}
		\dot{V}_\xi(\chi,\xi) &\leq - m\begin{bmatrix}
			\norm{\zeta(\x,\uu,\rr)}
			\\
			\norm{\xi}
		\end{bmatrix}^2.\label{eq:dot_V_m}
	\end{align}
	Let $\cU \coloneqq \{\uu \in \R^m \mid \nabla \fsh(\uu) = 0\} \subset \R^m$.
	Then, the above inequality allows us to guarantee that $\dot{V}(\chi) < 0$ for any $(\chi,\xi) \in E \times \R^{N}$, where the set $E \subset \R^{\nchi}$ reads as
	\begin{align}
		E  &\coloneqq  \{\chi  =   \col(\x,\uu,\rr)  \in  \R^{\nchi} \mid \x  =   h(\uu),  \uu  \in  \mathcal{U}, \rr  =  0\}.
	\end{align} 
	We thus achieved an extension of the claim obtained in the proof of~\cite[Th.~3.1]{carnevale2024nonconvex}, since here also the variable $\xi$ has been handled.
	Hence, the first two claims of the theorem follow by resorting to the same arguments in the conclusion of the proof of~\cite[Th.~3.1]{carnevale2024nonconvex}.

	As for the exclusion of the Zeno behavior, we proceed by contradiction.
	Suppose, without loss of generality, that an agent $i$ exhibits the Zeno behavior, namely 
	\begin{align}\label{eq:zeno_lim}
		\lim_{k_i \to \infty} \tki = t_i^\infty,
	\end{align}
	for some $t_i^\infty > 0$.
	The inequality~\eqref{eq:dot_V_m} proves the invariance of the level sets of $V_{\chi}$ for system~\eqref{eq:perturbed}.
	Further, since is $V(\chi)$ radially unbounded~\cite{carnevale2024nonconvex}, $V_\chi$ is radially unbounded too and, thus, its level sets are compact.
	This claim, combined with the definition of each $\ei$ (cf.~\eqref{eq:e_i}) and the Lipschitz properties assumed in Assumption~\ref{ass:steady_state} and~\ref{ass:lipschitz}, allows us to guarantee the existence of $M > 0$ such that
	\begin{align}
		\tfrac{d}{dt}\norm{\ei(t)} \leq M,\label{eq:bound_c_e}
	\end{align}
	for all $t \ge 0$ and $i \in \until{N}$.
	Since the protocol~\eqref{eq:triggering_law} imposes $\ei(t) = 0$ at the beginning of each slot $[\tki,\tkpi)$, the bound~\eqref{eq:bound_c_e} leads to
	\begin{align}
		\ei(t) = \ei(\tki) + \int_{\tki}^t
		\dfrac{d\norm{\ei(\tau)}}{d\tau}d\tau \leq M(t-\tki).\label{eq:bound_e_i_final}
	\end{align}
	On the other hand, by~\eqref{eq:xi}, it holds $\xi_i(t) = \xi_i(0)\exp(-\nu t)$ for all $t \ge 0$. 
	Thus, being $\tr\norm{\diri(\x_i,\ui,\w_i,\z_i)} \ge 0$ for any $t \ge 0$, the bound in~\eqref{eq:bound_e_i_final} imposes, as a necessary condition to satisfy the triggering in~\eqref{eq:triggering_law}, that
	\begin{align}\label{eq:triggering_law_tilde}
		M(\tkpi - \tki) \ge |\xi_i(0)|\exp(-\nu \tkpi)
	\end{align}
	From~\eqref{eq:zeno_lim}, for all $\epsilon>0$ there exists $k_{i,\epsilon} \in \N$ such that
	\begin{align}\label{eq:condition_limit}
		\tki \in [t_i^\infty - \epsilon, t_i^\infty], \quad \forall k_i \ge k_{i,\epsilon}.
	\end{align}
	Set
	\begin{align}
		\epsilon \coloneqq \frac{|\xi_i(0)|}{2M}\exp(-\nu t_i^\infty),
		\label{eq:epsilon}
	\end{align}
	and suppose that the $k_{i,\epsilon}$-th triggering time of agent $i$, namely $t_{i}^{k_{i,\epsilon}}$, has occurred. 
	Let $t_i^{k_{i,\epsilon} + 1}$ be the next triggering time determined by~\eqref{eq:triggering_law}. 
	Then, by using~\eqref{eq:triggering_law_tilde}, we get
	\begin{align}
		t_i^{k_{i,\epsilon} + 1} - t_i^{k_{i,\epsilon}} &\ge %
		\frac{|\xi_i(0)|}{M}\exp(-\nu t_i^{k_{i,\epsilon} + 1})
		\notag\\
		&\stackrel{(a)}{\geq} 
		\frac{|\xi_i(0)|}{M}\exp(-\nu t_i^\infty)
		\stackrel{(b)}{=} 
		2\epsilon,
		\label{eq:inequality_epsilon}
	\end{align} 
	where in $(a)$ %
	we use $t_i^\infty \ge t_i^{k_{i,\epsilon} + 1}$, while in $(b)$ we use~\eqref{eq:epsilon}.
	However, the inequality~\eqref{eq:inequality_epsilon} implies
	\begin{align*}
		t_i^{k_{i,\epsilon}} \leq t_i^{k_{i,\epsilon} + 1} - 2\epsilon \leq t_i^\infty - 2\epsilon,
	\end{align*}
	which contradicts~\eqref{eq:condition_limit} and concludes the proof.
	
	\bibliographystyle{IEEEtran}
	\bibliography{biblio}

\end{document}